\documentclass{article}
\usepackage[dvipsnames,table]{xcolor}

\usepackage{microtype}
\usepackage{graphicx}
\usepackage{subcaption}
\usepackage{booktabs} 

\usepackage{hyperref}

\usepackage[preprint]{icml2026}

\usepackage{amsmath}
\usepackage{amssymb}
\usepackage{mathtools}
\usepackage{amsthm}

\usepackage[capitalize,noabbrev]{cleveref}

\theoremstyle{plain}

\theoremstyle{definition}

\theoremstyle{remark}

\usepackage[textsize=tiny]{todonotes}

\usepackage{hyperref}
\usepackage{multirow}
\usepackage{xcolor}

\usepackage[dvipsnames,table]{xcolor}

\usepackage{mdframed}
\usepackage{listings}

\usepackage[most]{tcolorbox}

\tcbuselibrary{skins,breakable}
\tcbset{draft=false}
\usepackage{pifont}

\definecolor{TrajPurple}{HTML}{6F5BD6}   
\definecolor{TrajGrayFrame}{HTML}{DADAE3}
\definecolor{TrajGrayBack}{HTML}{F6F6F9}
\definecolor{TrajInk}{HTML}{2F2F35}

\newtcolorbox{Trajectory}[2][]{%
  enhanced,
  breakable,
  colback=white,
  colframe=TrajGrayFrame,
  boxrule=0.7pt,
  arc=3mm,
  left=2mm,right=2mm,top=1.5mm,bottom=2mm,
  title={#2},
  coltitle=TrajInk,
  fonttitle=\bfseries,
  boxed title style={
    colback=TrajGrayBack,
    colframe=TrajGrayFrame,
    arc=3mm,
    left=3mm,right=3mm,top=1.3mm,bottom=1.3mm,
    boxrule=0.7pt,
    yshift=-1.2mm
  },
  #1
}

\newtcolorbox{TrajAgentCard}[2]{%
  enhanced,
  breakable,
  colback=TrajPurple!6,
  colframe=TrajPurple!55!TrajGrayFrame,
  boxrule=0.65pt,
  arc=2.5mm,
  left=3mm,right=3mm,top=2.2mm,bottom=2.2mm,
  before skip=3mm, after skip=0mm,
  title=\strut #1\hfill{\footnotesize\color{TrajInk!55}#2},
  fonttitle=\bfseries,
  coltitle=TrajInk,
  boxed title style={
    colback=white,
    colframe=TrajPurple!55!TrajGrayFrame,
    arc=2.5mm,
    left=2.5mm,right=2.5mm,top=1.1mm,bottom=1.1mm,
    boxrule=0.65pt,
    yshift=-1.1mm
  }
}

\newtcolorbox{TrajGrayCard}[2]{%
  enhanced,
  breakable,
  colback=TrajGrayBack,
  colframe=TrajGrayFrame,
  boxrule=0.65pt,
  arc=2.5mm,
  left=3mm,right=3mm,top=2.2mm,bottom=2.2mm,
  before skip=3mm, after skip=0mm,
  title=\strut #1\hfill{\footnotesize\color{TrajInk!55}#2},
  fonttitle=\bfseries,
  coltitle=TrajInk,
  boxed title style={
    colback=white,
    colframe=TrajGrayFrame,
    arc=2.5mm,
    left=2.5mm,right=2.5mm,top=1.1mm,bottom=1.1mm,
    boxrule=0.65pt,
    yshift=-1.1mm
  }
}

\newtcbox{\codebox}{on line,
  colback=white,
  colframe=TrajGrayFrame,
  boxrule=0.45pt,
  arc=1.2mm,
  left=1.2mm,right=1.2mm,top=0.6mm,bottom=0.6mm,
  fontupper=\ttfamily\small\color{TrajInk}
}

\providecommand{\faRobot}{}
\providecommand{\faWrench}{}

\newcommand{\TrajAgent}{\faRobot\quad Agent}
\newcommand{\TrajTool}{\faWrench\quad Tool}

\newcounter{trajstep}

\newtcolorbox{TrajAgentStepCard}[1]{%
  enhanced,
  breakable,
  colback=TrajPurple!6,
  colframe=TrajPurple!55!TrajGrayFrame,
  boxrule=0.65pt,
  arc=2.5mm,
  left=3mm,right=3mm,top=2.2mm,bottom=2.2mm,
  before skip=3mm, after skip=0mm,
  title=\strut \TrajAgent\quad {\color{TrajInk!55}Step \#\refstepcounter{trajstep}\thetrajstep}\hfill{\footnotesize\color{TrajInk!55}#1},
  fonttitle=\bfseries,
  coltitle=TrajInk,
  boxed title style={
    colback=white,
    colframe=TrajPurple!55!TrajGrayFrame,
    arc=2.5mm,
    left=2.5mm,right=2.5mm,top=1.1mm,bottom=1.1mm,
    boxrule=0.65pt,
    yshift=-1.1mm
  }
}

\newtcolorbox{TrajToolStepCard}[2]{%
  enhanced,
  breakable,
  colback=TrajGrayBack,
  colframe=TrajGrayFrame,
  boxrule=0.65pt,
  arc=2.5mm,
  left=3mm,right=3mm,top=2.2mm,bottom=2.2mm,
  before skip=3mm, after skip=0mm,
  title=\strut \TrajTool\ \texttt{#1}\quad {\color{TrajInk!55}Step \#\refstepcounter{trajstep}\thetrajstep}\hfill{\footnotesize\color{TrajInk!55}#2},
  fonttitle=\bfseries,
  coltitle=TrajInk,
  boxed title style={
    colback=white,
    colframe=TrajGrayFrame,
    arc=2.5mm,
    left=2.5mm,right=2.5mm,top=1.1mm,bottom=1.1mm,
    boxrule=0.65pt,
    yshift=-1.1mm
  }
}

\usepackage[normalem]{ulem}
\usepackage{subcaption} 

\usepackage[textsize=tiny]{todonotes}

\usepackage{enumitem}

\definecolor{headerpurple}{RGB}{90, 70, 110}
\definecolor{contentbg}{RGB}{252, 250, 255}
\definecolor{highlightpurple}{RGB}{120, 60, 140}
\definecolor{bordercolor}{RGB}{140, 120, 160}

\newtcolorbox{toolbox}[2][]{%
  enhanced,
  arc=3mm,
  boxrule=0.5pt,
  colframe=bordercolor,
  colback=contentbg,
  width=\linewidth,
  left=2.5mm,
  right=2.5mm,
  top=2mm,
  bottom=2mm,
  title={#2},
  fonttitle=\bfseries\small\sffamily,
  colbacktitle=headerpurple,
  coltitle=white,
  toptitle=1.2mm,
  bottomtitle=1.2mm,
  fontupper=\small\ttfamily,
  before upper={\raggedright\setlength{\parindent}{0pt}},
  #1
}

\icmltitlerunning{SkillCraft: Can LLM Agents Learn to Use Tools Skillfully?}

\newcommand{\auth}[2]{\textbf{#1}\textsuperscript{#2}}

\begin{document}

\twocolumn[
\icmltitle{SkillCraft: Can LLM Agents Learn to Use Tools Skillfully?}

  \begin{center}
    \auth{Shiqi Chen}{1*}\quad
    \auth{Jingze Gai}{2*}\quad
    \auth{Ruochen Zhou}{2*}\quad
    \auth{Jinghan Zhang}{3}\quad
    \auth{Tongyao Zhu}{5}\quad
    \auth{Junlong Li}{3}\\
    \auth{Kangrui Wang}{4}\quad
    \auth{Zihan Wang}{4}\quad
    \auth{Zhengyu Chen}{6}\quad
    \auth{Klara Kaleb}{1}\quad
    \auth{Ning Miao}{2}\\
    \auth{Siyang Gao}{2}\quad
    \auth{Cong Lu}{6}\quad
    \auth{Manling Li}{4}\quad
    \auth{Junxian He}{3}\quad
    \auth{Yee Whye Teh}{1}

    \vspace{4pt}

    \textsuperscript{1}University of Oxford\quad
    \textsuperscript{2}City University of Hong Kong\quad
    \textsuperscript{3}Hong Kong University of Science and Technology\\
    \textsuperscript{4}Northwestern University\quad
    \textsuperscript{5}National University of Singapore\quad
    \textsuperscript{6}Independent

    \vspace{4pt}

    {\textbf{Code:} \href{https://github.com/shiqichen17/SkillCraft}{\texttt{github.com/shiqichen17/SkillCraft}}} \\
    {\textbf{Webpage:} \href{https://skillcraft-website.github.io/page}{\texttt{skillcraft-website.github.io/page}}}
  \end{center}

  \icmlkeywords{Machine Learning, ICML}

  \vskip 0.3in
]

\makeatletter
\gdef\icmlcorrespondingauthor@text{%
  \begingroup
  \footnotesize
  \setlength{\tabcolsep}{0pt}%
  \renewcommand{\arraystretch}{1.05}%
  \begin{tabular}[t]{@{}l@{\hspace{0.25em}}p{0.80\columnwidth}@{}}
    \multicolumn{2}{@{}l@{}}{\textsuperscript{*}Equal Contribution} \\
  \end{tabular}%
  \endgroup
}
\renewcommand{\printAffiliationsAndNotice}[1]{\global\icml@noticeprintedtrue%
  \stepcounter{@affiliationcounter}%
  {\let\thefootnote\relax\footnotetext{\hspace*{-\footnotesep}\ificmlshowauthors #1\fi%
      \forloop{@affilnum}{1}{\value{@affilnum} < \value{@affiliationcounter}}{%
        \textsuperscript{\arabic{@affilnum}}\ifcsname @affilname\the@affilnum\endcsname%
          \csname @affilname\the@affilnum\endcsname%
        \else
          {\bf AUTHORERR: Missing \textbackslash{}icmlaffiliation.}
        \fi
      }
      \ifdefined\icmlcorrespondingauthor@text
         \icmlcorrespondingauthor@text
      \else
        {\bf AUTHORERR: Missing \textbackslash{}icmlcorrespondingauthor.}
      \fi
      \\
      \Notice@String
    }%
  }%
}
\makeatother

\printAffiliationsAndNotice{}  

\definecolor{gold}{RGB}{205,133,63}
\definecolor{fGreen}{RGB}{34,139,34}
\definecolor{tOrange}{RGB}{255,207,151}
\definecolor{tBlue}{RGB}{165,238,255}
\definecolor{lightblue}{RGB}{224,235,255}
\definecolor{darkgray}{RGB}{181,186,186}
\definecolor{tPurple}{RGB}{240,177,254}
\definecolor{tPink}{RGB}{254,189,208}
\definecolor{tGreen}{RGB}{204,255,180}
\definecolor{tGold}{RGB}{255,215,0}
\definecolor{ood}{rgb}{0.95, 0.98, 1.0}
\newcommand{\up}[1]{\textcolor{OliveGreen}{\large \ $\uparrow${#1}}}
\newcommand{\down}[1]{\textcolor{Maroon}{\large \ $\downarrow${#1}}}

\newcommand{\upnew}[1]{\textcolor{OliveGreen}{\small \ $\uparrow${#1}}}
\newcommand{\downnew}[1]{\textcolor{Maroon}{\small \ $\downarrow${#1}}}

\newcommand{\downbad}[1]{\textcolor{Maroon}{\large \ $\downarrow${#1}}}

\newcommand{\shiqi}[1]{\textcolor{magenta}{$_{shiqi}$[#1]}}
\newcommand{\congl}[1]{\textcolor{blue}{$_{cong}$[#1]}}

\definecolor{TPBlueFrame}{HTML}{9CA3AF}
\definecolor{TPBlueBack}{HTML}{F3F4F6}
\definecolor{TPTitleBack}{HTML}{D1D5DB}
\definecolor{TPInk}{HTML}{1F2937}

\newcommand{\tperr}[1]{\textcolor{red!70!black}{#1}}

\newtcolorbox{TaskPromptBox}[1]{%
  enhanced,
  breakable,
  colback=TPBlueBack,
  colframe=TPBlueFrame,
  boxrule=0.7pt,
  arc=2.2mm,
  left=3mm,right=3mm,top=2.2mm,bottom=2.4mm,
  title={#1},
  fonttitle=\bfseries,
  coltitle=TPInk,
  boxed title style={
    colback=TPTitleBack,
    colframe=TPBlueFrame,
    arc=2.2mm,
    left=3mm,right=3mm,top=1.1mm,bottom=1.1mm,
    boxrule=0.7pt,
    yshift=-0.8mm
  }
}

\newcommand{\TPLabel}[1]{\textbf{\color{TPInk}#1}\hspace{0.6em}}

\newcommand{\TPDivider}{\par\noindent\textcolor{TPBlueFrame!70}{\rule{\linewidth}{0.4pt}}\par}

\definecolor{SPGrayFrame}{HTML}{9CA3AF}
\definecolor{SPGrayBack}{HTML}{F3F4F6}
\definecolor{SPTitleBack}{HTML}{D1D5DB}
\definecolor{SPInk}{HTML}{1F2937}

\newtcolorbox{SystemPromptBox}[1]{%
  enhanced,
  breakable,
  colback=SPGrayBack,
  colframe=SPGrayFrame,
  boxrule=0.7pt,
  arc=2.2mm,
  left=3mm,right=3mm,top=2.2mm,bottom=2.4mm,
  title={#1},
  fonttitle=\bfseries,
  coltitle=SPInk,
  boxed title style={
    colback=SPTitleBack,
    colframe=SPGrayFrame,
    arc=2.2mm,
    left=3mm,right=3mm,top=1.1mm,bottom=1.1mm,
    boxrule=0.7pt,
    yshift=-0.8mm
  }
}

\newcommand{\SPLabel}[1]{\textbf{\color{SPInk}#1}\hspace{0.6em}}

\newcommand{\SPDivider}{\par\noindent\textcolor{SPGrayFrame!70}{\rule{\linewidth}{0.4pt}}\par}

\begin{abstract}

Real-world tool-using agents operate over long-horizon workflows with recurring structure and diverse demands, where effective behavior requires not only invoking atomic tools but also abstracting, and reusing higher-level tool compositions. However, existing benchmarks mainly measure instance-level success under static tool sets, offering limited insight into agents’ ability to acquire such reusable skills.
We address this gap by introducing \textbf{SkillCraft}, a benchmark explicitly stress-test agent ability to form and reuse higher-level tool compositions, where we call \emph{Skills}. SkillCraft features realistic, highly compositional tool-use scenarios with difficulty scaled along both quantitative and structural dimensions, designed to elicit skill abstraction and cross-task reuse. We further propose a lightweight evaluation protocol that enables agents to auto-compose atomic tools into executable Skills, cache and reuse them inside and across tasks, thereby improving efficiency while accumulating a persistent library of reusable skills.
Evaluating state-of-the-art agents on SkillCraft, we observe substantial efficiency gains, with token usage reduced by up to 80\% by skill saving and reuse. Moreover, success rate strongly correlates with tool composition ability at test time, underscoring compositional skill acquisition as a core capability.

\begin{figure}[t!]
    \centering \includegraphics[width=\columnwidth]{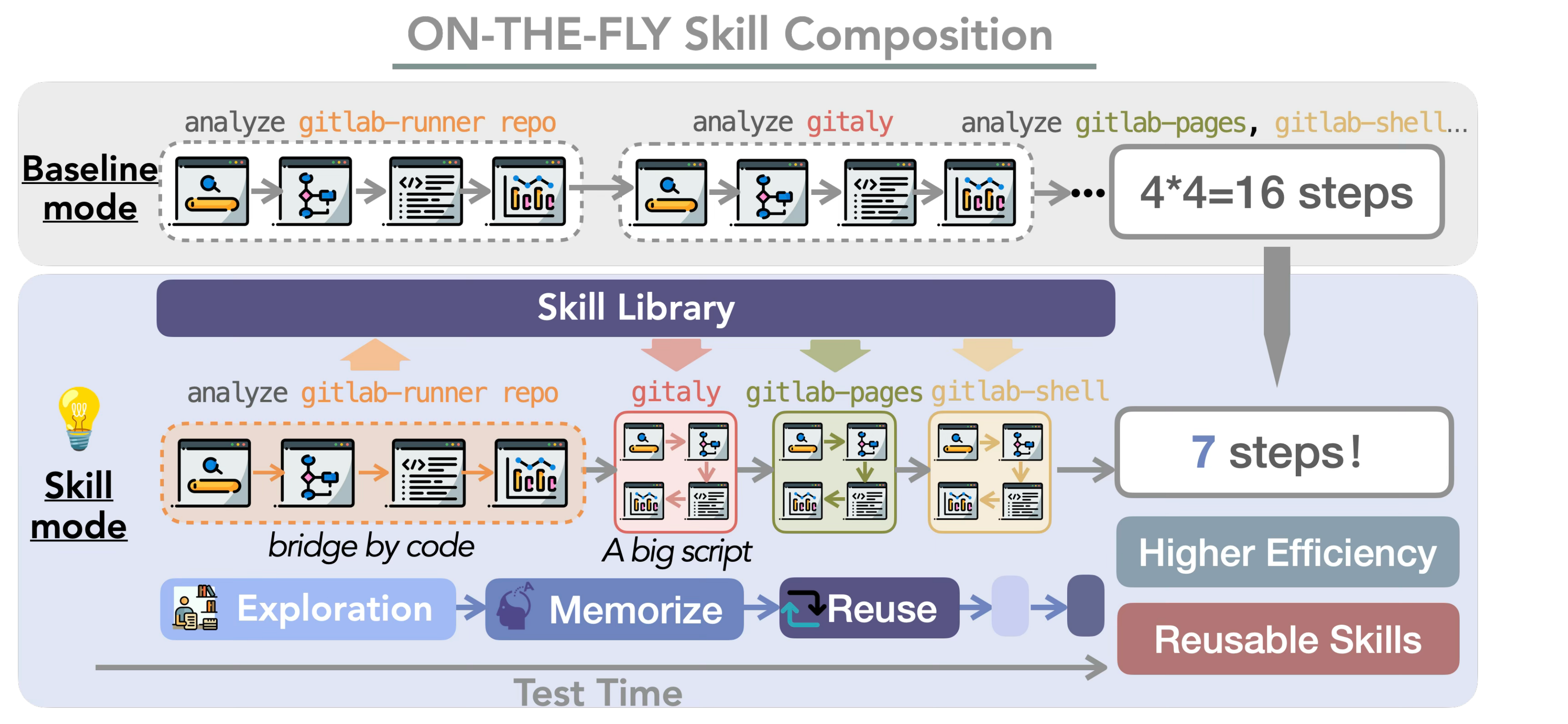} 
    \caption{Skill Mode demo. Demonstrating how skills are automatically discovered, cached locally, and subsequently reused.}
    \label{fig:wide-figure}
\end{figure}
\end{abstract}
\section{Introduction}

\begin{figure*}[t!]
    \centering
    \includegraphics[width=\textwidth]{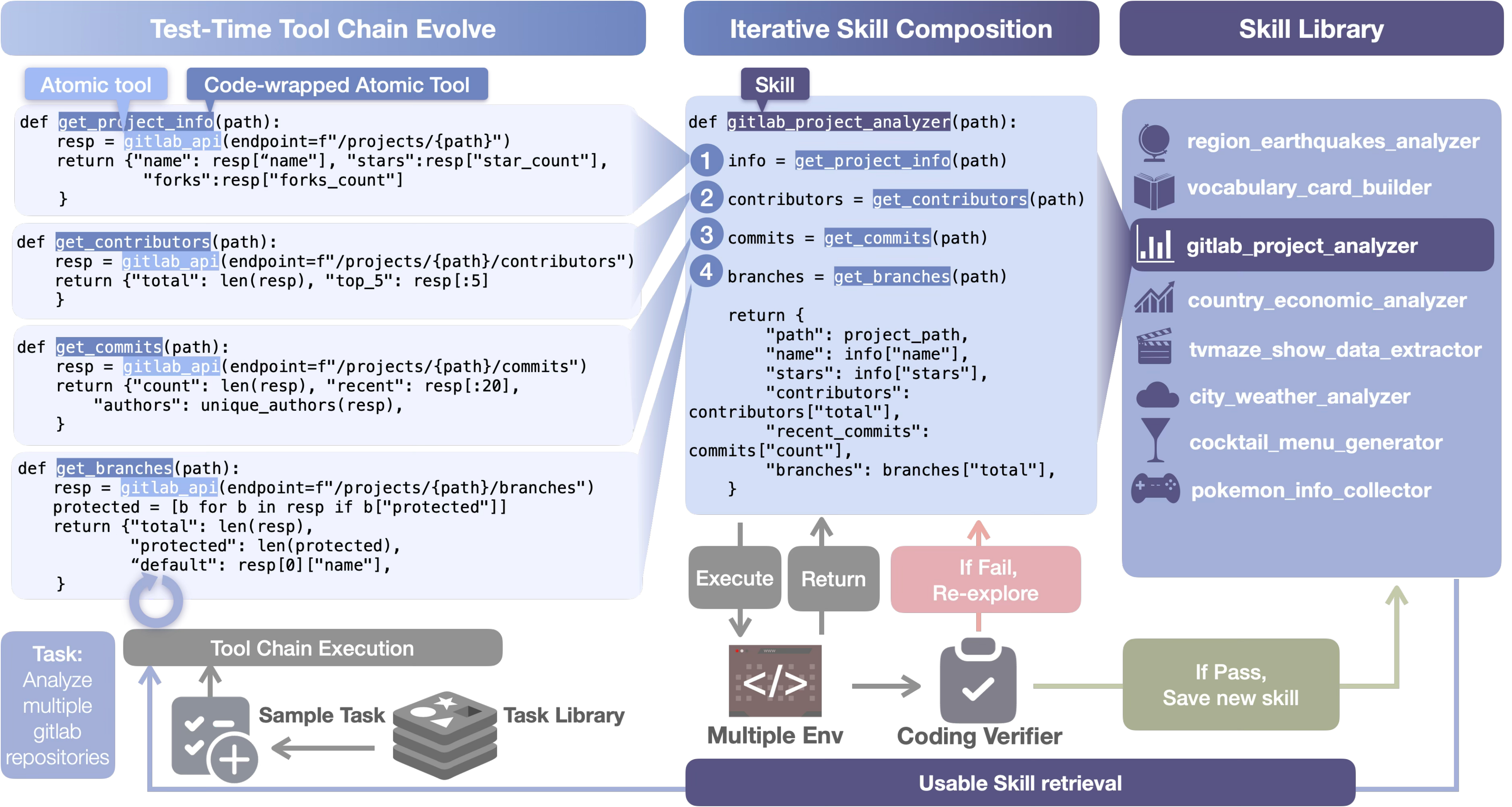} 
    \caption{\textbf{SkillCraft Protocol Pipeline Overview.}
The pipeline consists of three stages: 
\textbf{(1) Test-Time Tool-Chain Evolution:} The agent solves tasks from the Task Library by exploring and chaining atomic tools, forming executable tool sequences. 
\textbf{(2) Iterative Skill Composition:} Successful sequences are abstracted into candidate skills, executed and verified in a coding environment; failed executions trigger re-exploration, while validated skills are stored. 
\textbf{(3) Skill Library and Reuse:} A growing repository of verified, reusable skills that can be retrieved in later tasks to replace low-level tool exploration, enabling test-time skill accumulation and efficient composition.}

    \label{fig:wide-figure}
\end{figure*}


\begin{quote}
\textit{``The intelligence of a system is a measure of its skill-acquisition efficiency over a scope of tasks, with respect to priors, experience, and generalization difficulty.''} \\
\hfill -- Fran\c{c}ois Chollet, \textit{On the Measure of Intelligence}
\end{quote}

Real-world tool-using language agents increasingly operate in long-horizon workflows with recurring substructures, such as repeated search–analyze–summarize patterns across documents, repositories, or web services.~\citep{boisvert2024workarenacompositionalplanningreasoningbased,jimenez2024swebench,zhang2025swebenchgoeslive} In cognitive science, such repetition is precisely what gives rise to \emph{skill abstraction}: intelligence is characterized not by executing isolated actions, but by efficiently acquiring, reusing, and recomposing higher-level procedures from experience. In this view, effective behavior requires the ability to form \emph{compositional skills}, which are reusable tool compositions that capture shared structure across tasks rather than repeatedly solving each instance from scratch with flat, atomic tool calls. This raises a fundamental question: \emph{can an agent acquire and reuse such compositional tool skills that generalize across structurally similar tasks?}

Existing tool-using benchmarks~\citep{zhou2023webarena, xu2024theagentcompanybenchmarkingllmagents, li2025tool} typically \emph{fix} both the toolset and the model at deployment and adopt the paradigm: \textit{Can the agent solve this task with the given tools?} As a result, they provide limited signal on whether agents can accumulate, abstract, and reuse compositional skills across tasks. To isolate and measure this missing capability, we introduce \textbf{SkillCraft}, a benchmark with standardized protocols specifically designed to elicit and evaluate reusable tool compositions (Skills) within and across tasks. 
Unlike existing benchmarks, \textbf{SkillCraft} embeds repeated substructures within a single task, requiring agents to identify and reuse tool compositions multiple times within a fixed budget.

We construct \textbf{SkillCraft} in a three-stage manner. 
First, we explore existing tool-using tasks such as Toolathlon~\citep{li2025tool}, AgentCompany~\citep{xu2024theagentcompanybenchmarkingllmagents}, and WebArena~\citep{zhou2023webarena} to identify task design principles. 
Second, we construct seed tasks by both selecting and adapting high-quality tasks from existing benchmarks and carefully designing long-horizon tasks from scratch. 
Third, we scale task difficulty along two orthogonal dimensions to encourage tool composition and Skill abstraction.
\textbf{Quantitative scaling} increases the number of entities involved in a task. For example, a task is extended from ``analyze the commit history of repository A" to ``analyze five repositories", encouraging the reuse of learned Skills.
\textbf{Complexity scaling} links multiple subtasks into longer chains, increasing structural difficulty and enabling higher-level skill formation (e.g., fetching commits, identifying contributors, and correlating them).
These settings reflect realistic long-horizon tool use, where reusable high-level compositions are essential for efficient and robust problem solving.

In addition, we introduce a protocol to evaluate agents’ tool composition ability. We equip agents with a plug-and-play composition mechanism, termed~\textbf{Skill Mode}, which enables them to (i) automatically discover and cache successful sequences of tool calls as reusable skills, and (ii) invoke these cached skills on new inputs when similar patterns arise. In practice, we achieve this by modifying the system prompt and registering a set of tools that allow agents to save and execute Skills in a plug-and-play manner. This creates test-time tool evolution: agents expand their action space through discovery and reuse \emph{during the test time}, accumulating capabilities during solving tasks. 

Using \textbf{SkillCraft}, we evaluate state-of-the-art models (e.g., Gemini-2.5-Pro, Claude-Sonnet-4.5, GPT-5.1) and find that \emph{Skill Mode} substantially improves efficiency, reducing token usage by up to 80\%. Moreover, efficiency gains from tool composition strongly correlate with task success, indicating that stronger models are better at discovering, reusing, and exploiting recurring tool-use patterns under the same composition mechanism.
These results suggest that stronger models tend to benefit more from reusable tool compositions, and are better able to identify, reuse, and exploit recurring tool-use patterns under the same composition mechanism.

We further conduct a fine-grained analysis of composition quality along two complementary dimensions: \emph{depth} and \emph{generalization}. We find that deeper, automatically generated hierarchies are often not a reliable scaling strategy—despite high per-skill execution rates, nesting amplifies error propagation and debugging overhead—whereas well-tested, shallow skill libraries remain more robust and cost-effective. In contrast, truly high-quality compositions exhibit strong transferability: skills learned at one difficulty level can be statically reused at other levels (and even across models) with consistently high execution success, improving both success and efficiency.
\section{SkillCraft}


\begin{figure*}[t!]
    \centering
    \includegraphics[width=0.9\textwidth]{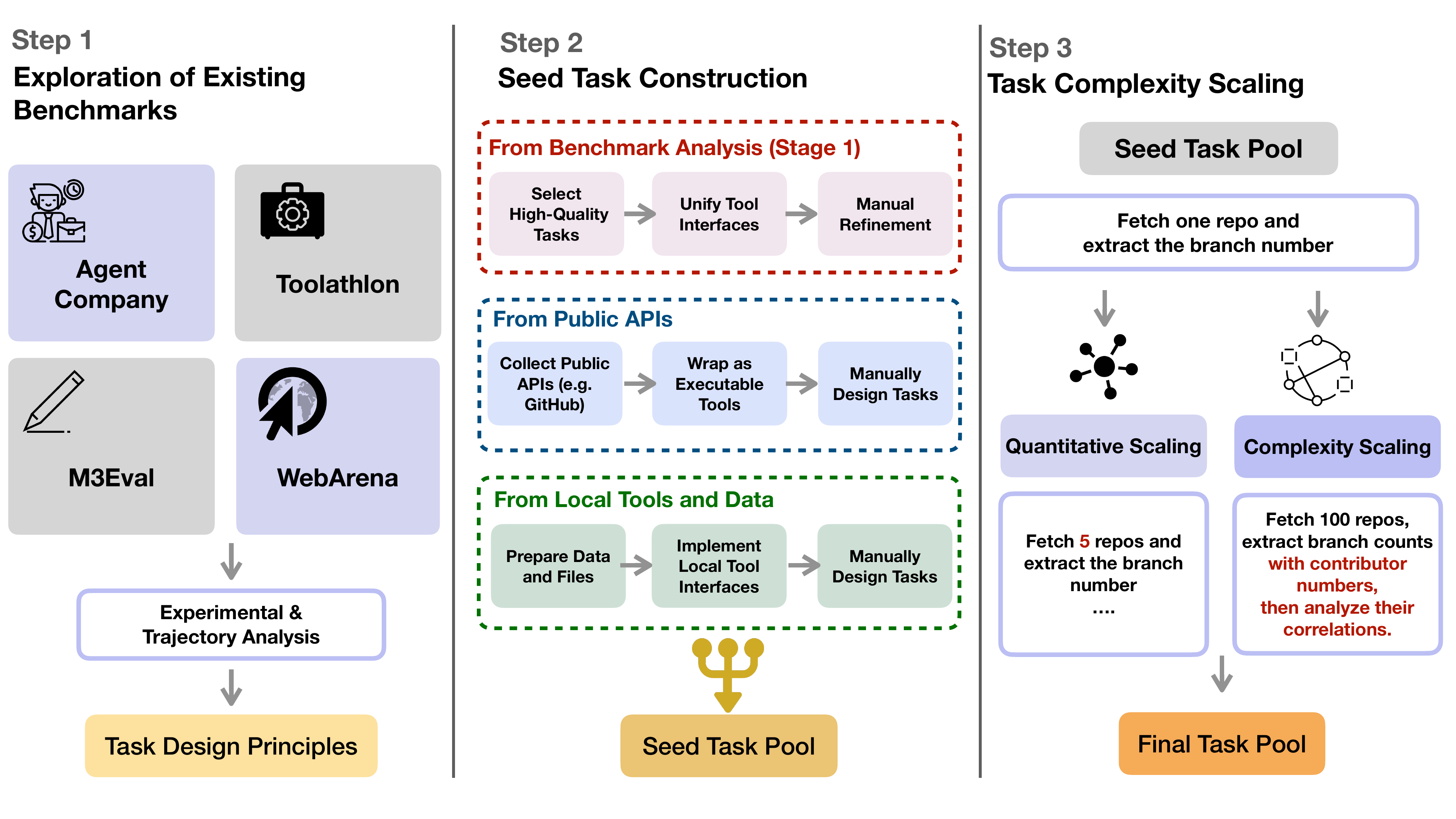}
    \caption{\textbf{Three-stage task construction pipeline for \textsc{SkillCraft}.} 
    In \textbf{Stage 1}, we explore existing benchmarks through systematic experimentation to identify effective \textbf{task design principles}. 
    In \textbf{Stage 2}, we construct seed tasks from three sources: (i) selected tasks from Stage 1 with \textbf{unified interfaces}, (ii) newly handcrafted \textbf{web API-based tasks}, and (iii) \textbf{local file and data processing tasks}. 
    In \textbf{Stage 3}, we systematically scale the seed tasks via \textbf{quantitative scaling} (increasing subtask count) and \textbf{complexity scaling} (increasing tool calls per subtask), producing a task repository with \textbf{graduated difficulty levels}.}
    \label{fig:wide-figure}
\end{figure*}


Current tool-using benchmarks mainly test whether agents can solve single task successfully with a fixed set of atomic tools (e.g., answering one real-time query with a search API). Such single-episode evaluations fail to reflect agents' tool composition ability. We therefore introduce \textbf{SkillCraft}, a long-horizon and compositional benchmark with repetitive structures that better reflects realistic settings and encourages the discovery and reuse of higher-level tool skills.




\subsection{What kinds of tasks can evaluate skill composition?}

We begin our exploration by asking: what kinds of tasks are required to faithfully evaluate an agent's ability to compose and reuse skills, rather than merely execute isolated tool calls?
To evaluate skill composition, tasks must go beyond single-shot, low-branching problems. If a task can be solved efficiently with a few atomic tool calls, agents have little incentive to discover or reuse higher-level skills, and composition ability becomes indistinguishable. We therefore seek tasks that resemble realistic workflows: they are long-horizon, structurally repetitive, and sufficiently challenging that solving them instance-by-instance is inefficient, making reusable tool compositions genuinely beneficial.

Guided by this motivation, our benchmark design follows two principles. First, tasks should require \emph{multi-step, multi-tool} reasoning, such that no single low-level tool call is sufficient and higher-level compositions provide a clear advantage. Second, tasks should exhibit \emph{recurrent structure with rich entity interactions} across instances, so that a skill discovered in one context can be meaningfully reused in others. This allows us to measure not only whether agents can compose atomic tools, but also whether the composed skills are reusable and generalizable. 

Importantly, these principles also mirror real-world tool-using scenarios, which are typically long-horizon and structurally repetitive, where similar sub-skills reoccur across tasks and the abstraction and reuse of higher-level skills are essential for efficient and robust problem solving.

\subsection{How to curate such tasks?}

\begin{figure}[t]
\centering
\includegraphics[width=0.8\columnwidth]{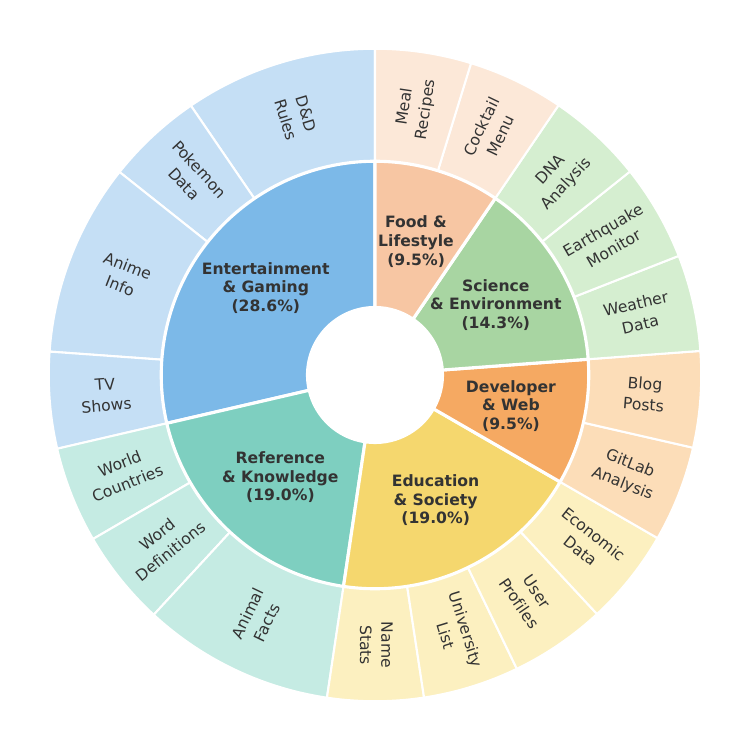}


\small
\renewcommand{\arraystretch}{1.05}
\setlength{\tabcolsep}{3pt}
\begin{tabular}{lcccc}
\toprule
\textbf{Difficulty} & \textbf{Tasks} & \textbf{Entity Num} & \textbf{Complexity} & \textbf{\%} \\
\midrule
Easy & 63 & 3 & 3 & 50.0\% \\
Medium & 42 & 4 & 4 & 33.3\% \\
Hard & 21 & 5 & 5 & 16.7\% \\
\midrule
\textbf{Total} & \textbf{126} & -- & -- & \textbf{100\%} \\
\bottomrule
\end{tabular}

\caption{Task distribution in SkillCraft. The chart shows 21 task families across 6 application domains. The table summarizes difficulty levels: ~\textbf{Entity Num} = number of target items (subtasks) per task;~\textbf{Complexity} = tool calls required per entity.}
\label{fig:task_distribution}
\end{figure}

We construct the benchmark through a three-stage pipeline. (1) \textbf{Exploratory Phase.} We first sample a set of complex, multi-step tool-using tasks from multiple existing agent benchmarks such as Toolathlon~\citep{li2025tool}, AgentCompany~\citep{xu2024theagentcompanybenchmarkingllmagents}, WebArena~\citep{zhou2023webarena} and M3ToolEval~\citep{wang2024executable}. 
Through systematic experimentation, we identify useful APIs\&task types and gain key insights that guide our task design principles. (2) \textbf{Seed Task Creation.} We construct our seed task pool from three sources: (i)a small set of high-quality tasks adapted from Stage 1 whose required APIs are reliable, stable, and free of severe rate limits, and whose difficulty is within the model’s capability, ensuring that large-scale, long-horizon interaction and tool composition are both feasible. (2)~\textbf{Seed Task Creation.} We build the seed task pool from three sources: (i) a small set of high-quality tasks adapted from Stage 1 with reliable, stable, and rate-limit–robust APIs; (ii) a large collection of handcrafted web API tasks; and (iii) local file and data processing tasks based on custom datasets. Stage-1 tasks are converted to a unified MCP interface. For web APIs, we survey, test, and filter stable public endpoints (e.g., GitLab, Open-Meteo, TVMaze), wrap them as standardized local tools, and design tasks accordingly. For local tasks, we prepare datasets, implement standardized processing tools, and construct tasks on top of them.
(3) \textbf{Systematic Scaling.} We expand seed tasks along two axes: (i) \textbf{quantitative scaling}, increasing the number of entities/subtasks, and (ii) \textbf{complexity scaling}, increasing tool calls per subtask. Combining the two yields multiple difficulty levels (e.g., $3\times3$, $4\times4$, $5\times5$), creating substantial headroom and encouraging discovery and reuse of higher-level compositional skills. Table~\ref{table:task-construction-stats} reports stage-wise statistics, and Fig~\ref{fig:task_distribution} shows coverage across domains and difficulty levels.


\begin{table*}[!h]
    \centering
    \setlength{\tabcolsep}{8pt}
    
    \caption{\textbf{Task statistics across the three-stage construction pipeline.} 
    In \textbf{Stage 1}, we explore \textbf{60+ tasks} from existing benchmarks to identify effective task design principles. 
    In \textbf{Stage 2}, we construct \textbf{21 seed tasks} from three sources: adapted benchmark tasks, handcrafted web API-based tasks, and local processing tasks. 
    In \textbf{Stage 3}, we systematically scale seed tasks by increasing \textbf{entity number} and \textbf{subtask complexity}, producing \textbf{126 tasks} across 6 difficulty levels.}
    
    \begin{tabular}{@{}p{4cm}p{6.5cm}p{3.5cm}c@{}}
        \toprule
        \textbf{Stage} & \textbf{Description} & \textbf{Source} & \textbf{\#Tasks} \\
        \midrule
        Stage 1: Exploratory Phase 
        & Explore existing benchmarks to identify task design principles 
        & Existing benchmarks like Toolathlon, WebArena, TextArena, M3Eval, etc.
        & 60+ \\
        \midrule
        Stage 2: Seed Task Creation & (i) Select \& adapt quality tasks from Stage 1 & Existing benchmarks & 5 \\
         & (ii) Handcraft web API-based tasks & GitLab, OpenMeteo, etc. & 12 \\
        & (iii) Handcraft local processing tasks & Custom datasets \& files & 4 \\
        
        \cmidrule(l){2-4}
        & \textbf{Total Seed Tasks} & & \textbf{21} \\
        \midrule
        Stage 3: Systematic Scaling & Scale entity number (N: 3→4→5) & From Seed tasks & \textbf{126} \\
         & Scale subtask complexity (M: 3→4→5) & Seed tasks × 6 levels & \\
        \bottomrule
    \end{tabular}

    \label{table:task-construction-stats}
\end{table*}

\subsection{How to Evaluate Tool Composition Ability?}

Inspired by cognitive science, which views intelligence as the efficiency of acquiring and reusing skills under limited resources~\citep{anderson1982acquisition,anderson1987skill,chollet2019measure}, we evaluate tool composition not only by task success but also by \emph{efficiency}. In our specific agentic tool-use setting, we also question whether efficiency remains a reliable evaluation metric. As a first step, we analyze the baseline setting to establish a reference point. Our analysis of current models operating with only low-level (atomic) tools reveals two recurring inefficiency patterns: (1) \textbf{Redundant state passing:} Intermediate results are repeatedly serialized between consecutive tool calls, incurring substantial token overhead.  (2) \textbf{Context window saturation:} Long sequences of tool calls and their outputs consume substantial context capacity, potentially causing the model to "forget" earlier information or lose track of the overall goal.

These observations expose a fundamental limitation: complex skills must be decomposed into sequences of atomic operations, each requiring explicit state passing and reasoning. A natural remedy is to \textbf{consolidate frequently co-occurring tool chains into a single executable unit}, which we term \emph{Skills}. Code provides a natural medium for this consolidation, compactly representing data flow, control logic, and iteration.

Accordingly, our evaluation asks: given multi-step, multi-tool tasks, can models abstract recurring tool chains into reusable, code-based Skills? Does this abstraction improve efficiency and success, as measured by \textbf{token usage}, \textbf{tool call count}, and \textbf{interaction steps}? We answer these questions by evaluating models on SkillCraft.
\section{SkillCraft Protocol}
In this section, we introduce the evaluation protocol for SkillCraft. To assess models' composition and skill curation abilities, we employ a pipeline that enables models to compose existing tools into novel higher-level ones and re-use them both inside current task and also cross-tasks. 
This evaluation protocol enables two core capabilities in a quantifiable process:
(1) Composition: Models could abstract multi-step tool chains into reusable code-based Skills.
(2) Reuse: Models retrieve and reuse the discovered Skills at test time, enabling graceful execution and accumulating efficiency gains over repeated interactions.

\subsection{Four Minimal MCP Primitives}
To support skill reuse with minimal system assumptions, we expose a lightweight MCP interface that allows an agent to store and reuse executable code-based Skills. In practice, we maintain a \emph{Skill Library} (a cache of \emph{verified} Skills and their metadata) and expose four lightweight MCP primitives as the only way to interact with this library.
This interface intentionally covers only the operational actions required by SkillMode: \emph{storage}, \emph{retrieval}, \emph{enumeration}, and \emph{execution}.
Specifically, the Skill Library is accessed through \texttt{save\_skill} (persist a workflow), \texttt{get\_skill} (retrieve code and metadata), \texttt{list\_skills} (discover available skills), and \texttt{execute\_skill} (run a skill as a higher-level tool).
Together, these primitives define the evaluation boundary: whether a model attempts reuse, whether reuse succeeds, and whether failures are handled can all be directly observed through these API calls. Figure~\ref{fig:macro-tools} illustrates the details about how these primitives fit into the overall protocol.

\subsection{Coding Verifier}
We introduce a Coding Verifier that applies three-stage validation before any Skill enters the library. The stages are: 

(a) Syntax Validation: Before accepting \texttt{save\_skill}, we parse the Skill code and reject syntactically invalid submissions, returning error line numbers and context snippets to block fundamentally broken code.

(b) Runtime Error Reporting: When \texttt{execute\_skill} fails, we return structured debugging information (e.g. exception messages, tracebacks, and input parameters), which enables models to distinguish syntax issues from tool invocation problems or parameter mismatches.

(c) Post-execution Quality Detection: To filter out useless Skills, we detect silent failures by checking output quality. For example, if over 50\% of output fields contain \emph{Unknown, None, or 0}, we flag the Skill as low-quality and reject it. 

\subsection{SkillCraft Protocol Pipeline}
To capture how models discover, store, and reuse skills across episodes, the protocol makes explicit, at each step, whether a previously learned skill can replace a sequence of atomic tool calls. The protocol proceeds as follows:





(1) \textbf{Reuse Attempt.} For new task, agent queries existing Skills by \texttt{list\_skills} and attempts to invoke a matching one by \texttt{execute\_skill} with task-specific parameters.

(2) \textbf{Exploration.} If no suitable Skill exists or execution fails, the agent solves the task with atomic tools and records the successful tool sequence.

(3) \textbf{Composition.} The successful sequence is abstracted into a parameterized candidate Skill, consolidating recurring subroutines and passing intermediate results through code variables rather than natural language.

(4) \textbf{Verification and Saving.} The candidate Skill is executed in a controlled \emph{Coding Env} via a unified \texttt{call\_tool()} interface and validated by a \emph{Coding Verifier}. Only skills that pass execution and verification are stored in the \emph{Skill Library} via \texttt{save\_skill} for reliable future reuse.


\section{Evaluation}



We evaluate agents on SkillCraft in a consistent and unified setting under the same task prompts, tool endpoints, and environment constraints. Here we introduce our settings.


\paragraph{Models} We benchmark a representative set of state-of-the-art models, including~\text{Kimi-K2-Thinking}~\citep{team2025kimi}, \text{DeepSeek-V3.2-EXP}~\citep{liu2025deepseek}, \text{DeepSeek-R1}~\citep{guo2025deepseek}, \text{Gemini-3-Pro}~\citep{googledeepmind2025gemini3pro_modelcard}, \text{Minimax-M2.1}\citep{minimax2025_m21}, \text{Claude-4.5-Sonnet}~\citep{anthropic2025_claude_sonnet_4_5_system_card} and \text{GPT-5.2}~\citep{openai2025_gpt52_system_card}.

\paragraph{Metrics}
We measure \textbf{Success Rate} using accuracy. For each task, we follow Toolathlon to define a human-expert, handcrafted evaluation rule for matching and scoring the outputs, counting a task as successful if its final score $\ge 90\%$. To measure Skill behavior beyond task completion, we report \textbf{Exec Rate}, the fraction of successful Skill executions among all Skill execution attempts, and \textbf{Reusing Rate}, the average number of times each saved Skill is invoked.

For efficiency metrics, we have \textbf{InTok/OutTok} (total input/output tokens) and \textbf{Turn Num} (LLM interaction rounds), and \textbf{Tool\_Call Num} when applicable. For each consumption metric $m$, we compute \textbf{Diff} as $(m_{\text{skill}}-m_{\text{base}})/m_{\text{base}}$ (negative indicates savings). To ensure fair comparisons, efficiency metrics are averaged over the subset of tasks where both compared modes succeed.

\newcommand{\better}[1]{\textcolor{green!60!black}{#1}}
\newcommand{\worse}[1]{\textcolor{red!70!black}{#1}}
\definecolor{baseblue}{RGB}{230, 240, 250}
\definecolor{skillgreen}{RGB}{230, 245, 235}
\begin{table*}[!h]\scriptsize
\centering
\Large
\renewcommand{\arraystretch}{1.4}
\setlength{\tabcolsep}{3.5pt}
\caption{
Results (base vs skill mode) across models on 126 tasks.
\textbf{Success Rate (Overall)}: task completion rate (score $\geq$ 90) for \colorbox{baseblue}{Baseline} (no skills) and \colorbox{skillgreen}{Skill} (with skills) modes, plus \textbf{Success Rate (Hard)} for the hard subset only.
\textbf{Skill Stats}: Exec = skill execution success rate; Reuse = average times each skill is invoked.
\textbf{Efficiency metrics} (Tokens, Cost, Turns, Tools): per-task averages computed over tasks where \emph{both} modes succeeded; each shows \colorbox{baseblue}{Base}, \colorbox{skillgreen}{Skill}, and Diff values.
\textbf{Diff}: percentage change (Skill $-$ Baseline) / Baseline; \better{negative} values indicate improvement, \worse{positive} values indicate degradation.
}
\resizebox{\textwidth}{!}{%
\begin{tabular}{l|cc|ccc|ccc|ccc|ccc|cc|cc}
\toprule
\multirow{2}{*}{\textbf{Model}} &
\multicolumn{2}{c|}{\textbf{Skill Stats}} &
\multicolumn{3}{c|}{\textbf{Avg Tokens}} &
\multicolumn{3}{c|}{\textbf{Avg Cost (\$)}} &
\multicolumn{3}{c|}{\textbf{Avg Turns}} &
\multicolumn{3}{c|}{\textbf{Avg Tool Calls}} &
\multicolumn{2}{c|}{\textbf{Success Rate (Overall)}} &
\multicolumn{2}{c}{\textbf{Success Rate (Hard)}} \\
\cmidrule{2-3}\cmidrule{4-6}\cmidrule{7-9}\cmidrule{10-12}\cmidrule{13-15}\cmidrule{16-17}\cmidrule{18-19}
& \textbf{Exec} & \textbf{Reuse} &
\cellcolor{baseblue}\textbf{Base} & \cellcolor{skillgreen}\textbf{Skill} & \textbf{Diff} &
\cellcolor{baseblue}\textbf{Base} & \cellcolor{skillgreen}\textbf{Skill} & \textbf{Diff} &
\cellcolor{baseblue}\textbf{Base} & \cellcolor{skillgreen}\textbf{Skill} & \textbf{Diff} &
\cellcolor{baseblue}\textbf{Base} & \cellcolor{skillgreen}\textbf{Skill} & \textbf{Diff} &
\cellcolor{baseblue}\textbf{Base} & \cellcolor{skillgreen}\textbf{Skill} &
\cellcolor{baseblue}\textbf{Base} & \cellcolor{skillgreen}\textbf{Skill} \\
\midrule
\multicolumn{19}{l}{\textit{\textbf{Open-Source Models}}} \\
\midrule
\textbf{Kimi-K2-Thinking} 
& 70\% & 3.4$\times$
& 0.51M & 0.30M & \better{-42\%}
& 0.21 & 0.13 & \better{-39\%}
& 6.7 & 8.3 & \worse{+24\%}
& 16.8 & 11.9 & \better{-29\%}
& 55/126 (44\%) & 56/126 (44\%)
& 8/21 (38\%) & 7/21 (33\%) \\
\textbf{DeepSeek-V3.2-EXP} 
& 71\% & 4.8$\times$ 
& 1.04M & 0.53M & \better{-49\%}
& 0.21 & 0.10 & \better{-51\%}
& 18.8 & 15.4 & \better{-18\%}
& 19.2 & 14.9 & \better{-23\%}
& 76/126 (60\%) & 87/126 (69\%) 
& 9/21 (42\%) & 15/21 (71\%) \\
\textbf{DeepSeek-R1} 
& 62\% & 3.4$\times$ 
& 0.58M & 0.41M & \better{-30\%}
& 0.24 & 0.18 & \better{-24\%}
& 9.0 & 9.9 & \worse{+10\%}
& 13.4 & 11.7 & \better{-12\%}
& 89/126 (71\%) & 101/126 (80\%) 
& 11/21 (52\%) & 15/21 (71\%) \\
\textbf{GLM-4.7} 
& 91\% & 3.7$\times$ 
& 0.78M & 0.48M & \better{-39\%}
& 0.20 & 0.12 & \better{-41\%}
& 13.5 & 13.0 & \better{-4\%}
& 16.9 & 13.3 & \better{-21\%}
& 91/126 (72\%) & 108/126 (86\%) 
& 12/21 (57\%) & 15/21 (71\%) \\
\textbf{Minimax-M2.1} 
& 100\% & 3.2$\times$ 
& 0.42M & 0.38M & \better{-11\%}
& 0.04 & 0.04 & \better{-8\%}
& 5.5 & 5.2 & \better{-6\%}
& 16.6 & 15.4 & \better{-7\%}
& 117/126 (93\%)  & 119/126 (94\%)
& 18/21 (86\%) & 20/21 (95\%) \\
\midrule
\multicolumn{19}{l}{\textit{\textbf{Closed-Source Models}}} \\
\midrule
\textbf{GPT-5.2} 
& 84\% & 3.8$\times$ 
& 1.23M & 0.26M & \better{-79\%}
& 1.77 & 0.43 & \better{-75\%}
& 20.6 & 9.9 & \better{-52\%}
& 19.4 & 8.9 & \better{-54\%}
& 109/126 (87\%) & 114/126 (90\%) 
& 16/21 (76\%) & 17/21 (80\%) \\
\textbf{Gemini 3 Pro} 
& 93\% & 3.9$\times$ 
& 0.66M & 0.30M & \better{-54\%}
& 0.59 & 0.30 & \better{-49\%}
& 10.5 & 11.9 & \worse{+13\%}
& 16.0 & 9.5 & \better{-41\%}
& 108/126 (86\%) & 116/126 (92\%) 
& 16/21 (76\%) & 17/19 (89\%) \\
\textbf{Claude 4.5 Sonnet} 
& 81\% & 3.4$\times$ 
& 1.36M & 0.40M & \better{-71\%}
& 1.08 & 0.28 & \better{-74\%}
& 15.3 & 10.2 & \better{-33\%}
& 14.3 & 9.2 & \better{-36\%}
& 119/126 (94\%) & 121/126 (96\%) 
& 20/21 (95\%) & 20/21 (95\%) \\
\bottomrule
\end{tabular}
}
\label{tab:main_res}
\end{table*}



\paragraph{Results} Table~\ref{tab:main_res} shows our main results. Overall,~\textbf{Skill Mode yields consistent and substantial gains in both success and efficiency across models}. For every model, Skill Mode sharply reduces average token usage and cost, and typically decreases the number of tool calls as well. However, the average number of conversation turns (highlighted in red in Table~\ref{tab:main_res} Avg Turns) can increase for some models, as Skill Mode adds extra decision and verification steps when selecting and executing cached skills. But these additional turns are typically lightweight, so overall tokens and cost still drop. For example, GPT-5.2 improves success from 109/126 (87\%) to 114/126 (90\%), and also cutting average tokens from 1.23M to 0.26M (-79\%) and average cost from \$1.77 to \$0.43 (-75\%). It suggests that once skills are discovered and cached, long-horizon tool-chain planning can be solved both more effectively and more efficiently through repeated reuse.

\begin{figure}[t!]
    \centering
    \includegraphics[width=\columnwidth]{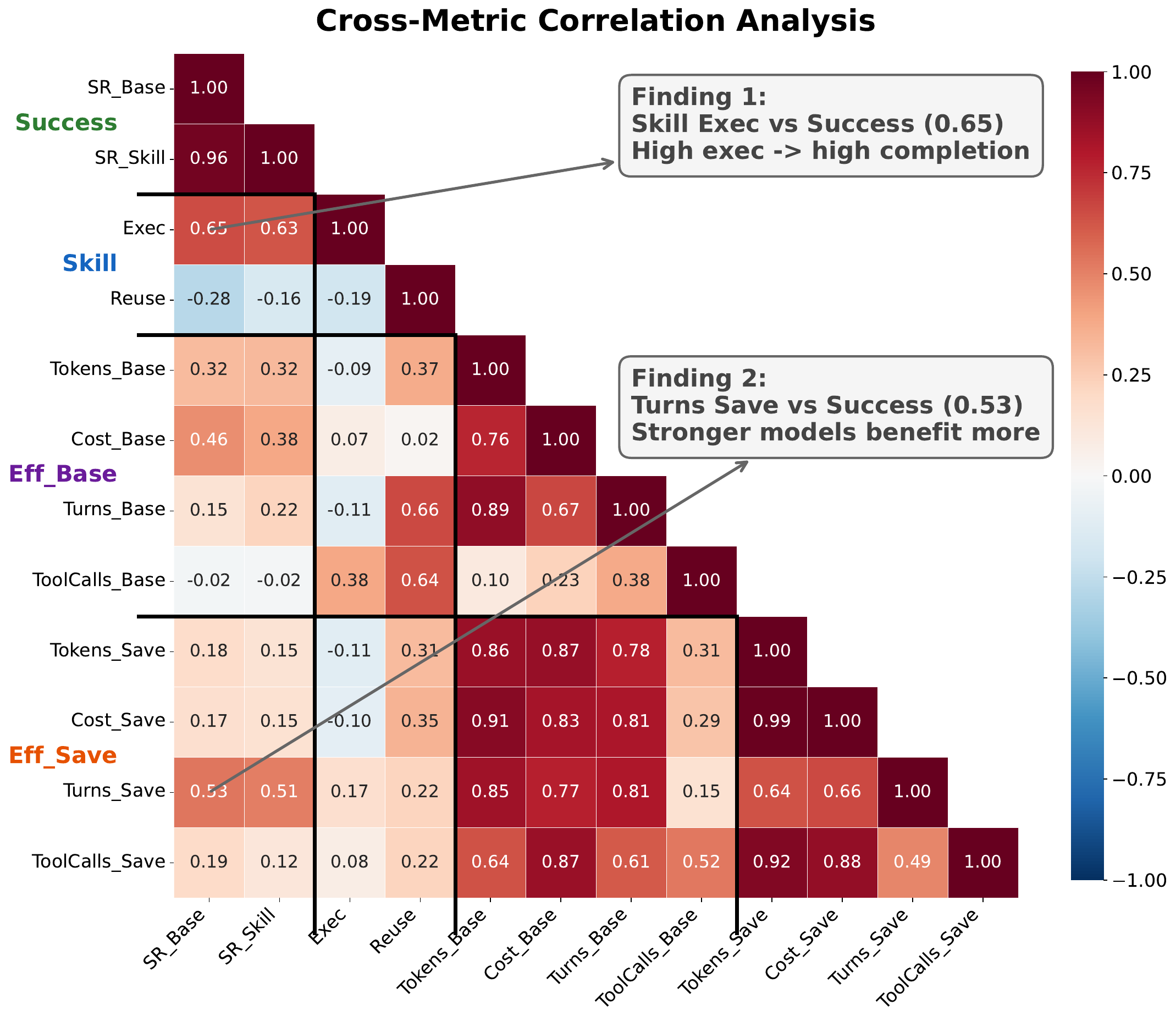} 
    \caption{Cross-metric correlation heatmap. Metrics are grouped into four categories: Success, Skill, Eff\_Base, and Eff\_Save. Key findings: (1) Skill execution rate correlates with task success (r=0.65); (2) Stronger models achieve greater efficiency gains from skills (r=0.53).}
\vspace{-10pt}
    \label{fig:heatmap}
\end{figure}

Moreover, ~\textbf{the magnitude of efficiency gains correlates positively with model capability}. Cross-metric correlation analysis shown in Figure~\ref{fig:heatmap}reveals two key patterns: (1)~\textbf{skill execution rate correlates with task success} (r=0.65), indicating that skill composition ability is tightly coupled with coding ability (skill execution success rate measures how reliably generated skills can be executed, with higher rates indicating better coding quality).; (2) \textbf{Efficiency savings correlate with baseline success} (e.g., $r=0.53$ for \emph{Turns Saved} and \emph{success rate}), confirming that stronger models benefit more from skill reuse. Concretely, \emph{closed-source} models such as Claude Sonnet 4.5 and GPT-5.2---which start from high baseline success (94\% and 87\%)---achieve the largest token reductions (-71\% and -79\%). In contrast, \emph{open-weight} models either suffer from lower success rates (\textless 90\% overall and \textless 60\% on the hard set; see Table~\ref{tab:difficulty_breakdown}), as observed for Kimi, DeepSeek, and GLM models, or exhibit limited tool-composition gains. For example, MiniMax-M2.1 shows only modest savings (-11\%), likely because it already solves many tasks efficiently without invoking skills. These findings suggest Skill Mode acts as a capability amplifier, benefiting models that can both synthesize correct skills and execute them reliably.

Moreover, our case studies reveal clear differences in tool composition behavior across models. Stronger models compose tools flexibly, invoking and reusing skills only when beneficial, whereas weaker models tend to follow prompts more rigidly and over-apply composition even when it is unnecessary. This supports the view that tool composition ability is a core metric of intelligence. Detailed examples are provided in the Appendix~\ref{app:traj}.

\section{What is a good tool composition?}
To understand what constitutes a \emph{good} tool composition, we study tool composition along two key dimensions: \textbf{composition depth} and \textbf{generalization ability}. Specifically, we examine whether deeper, hierarchical compositions lead to better performance, and whether learned Skills can generalize across tasks, complexity levels, and models.
\subsection{Is Deeper Composition Always Better?}
We introduce \textbf{Hierarchical Mode}, which enables hierarchical, tree-structured skill composition by allowing skills to invoke other skills during execution. Under the standard \textbf{SkillCraft} protocol (Skill Mode), supporting single-level composition: each skill is defined as a composition of atomic tool calls and cannot invoke other skills. Iteration Mode lifts this restriction by enabling recursive skill invocation, permitting hierarchical composition up to a configurable nesting depth (\textit{max\_skills\_nesting\_depth=10} in our experiments). In theory, hierarchical composition enables reusable abstraction, yields multiplicative efficiency gains through nested skill reuse, and allows the agent to reason at higher levels rather than managing low-level details.

In practice, under \textbf{Hierarchical Mode}, when a skill is executed, the \texttt{call\_tool} interface—responsible for dispatching executable actions during skill execution—can invoke not only atomic tools but also previously saved skills via \texttt{execute\_pattern}. In contrast, under the standard \textbf{SkillCraft} protocol, \texttt{call\_tool} is restricted to atomic tool invocations and cannot trigger other skills. This enables hierarchical/recursive skill invocation and yields a tree-structured execution graph, in which high-level skills orchestrate lower-level ones, as illustrated in Figure~\ref{fig:cases}(a).


\definecolor{baseblue}{RGB}{230, 240, 250}
\definecolor{skillgreen}{RGB}{230, 245, 235}
\definecolor{iteratepurple}{RGB}{240, 230, 250}

\begin{table}[!h]
\centering
\scriptsize
\renewcommand{\arraystretch}{1.1}
\setlength{\tabcolsep}{1.5pt}

\caption{Three-mode comparison across models. \textbf{Base}: No skill library. \textbf{Skill}: With skill library. \textbf{Hier}: Hierarchical mode with skill nesting. \textbf{N}: Number of successful tasks out of 126 total. \textbf{Ex}: Execution success rate (\%). \textbf{Re}: Reuse factor ($\times$). \textbf{$\Delta$}: Relative change vs.\ Base (\%).}

\resizebox{\columnwidth}{!}{%
\begin{tabular}{l|l|cc|cc|cc|cc|cc|cc}
\toprule
\multirow{2}{*}{\textbf{Model}} &
\multirow{2}{*}{\textbf{Mode}} &
\multicolumn{2}{c|}{\textbf{Success}} &
\multicolumn{2}{c|}{\textbf{Skill}} &
\multicolumn{2}{c|}{\textbf{Tokens}} &
\multicolumn{2}{c|}{\textbf{Cost (\$)}} &
\multicolumn{2}{c|}{\textbf{Turns}} &
\multicolumn{2}{c}{\textbf{Tools}} \\
\cmidrule{3-4}\cmidrule{5-6}\cmidrule{7-8}\cmidrule{9-10}\cmidrule{11-12}\cmidrule{13-14}
& & \textbf{N} & \textbf{\%} & \textbf{Ex} & \textbf{Re} & \textbf{Val} & \textbf{$\Delta$} & \textbf{Val} & \textbf{$\Delta$} & \textbf{Val} & \textbf{$\Delta$} & \textbf{Val} & \textbf{$\Delta$} \\
\midrule

\multirow{3}{*}{\textbf{DeepSeek-V3.2}}
& \cellcolor{baseblue}Base
& 76 & 60 & -- & -- & 1.04M & -- & 0.21 & -- & 18.8 & -- & 19.2 & -- \\
& \cellcolor{skillgreen}Skill
& 87 & 69 & 71 & 4.8 & 0.53M & \better{-49} & 0.10 & \better{-51} & 15.4 & \better{-18} & 14.9 & \better{-23} \\
& \cellcolor{iteratepurple}Hier
& 92 & 73 & 75 & 3.0 & 0.68M & \better{-36} & 0.10 & \better{-66} & 16.5 & \better{-13} & 15.4 & \better{-20} \\
\midrule

\multirow{3}{*}{\textbf{Claude-4.5}}
& \cellcolor{baseblue}Base
& 119 & 94 & -- & -- & 1.36M & -- & 1.08 & -- & 15.3 & -- & 14.3 & -- \\
& \cellcolor{skillgreen}Skill
& 121 & 96 & 81 & 3.4 & 0.40M & \better{-71} & 0.28 & \better{-74} & 10.2 & \better{-33} & 9.2 & \better{-36} \\
& \cellcolor{iteratepurple}Hier
& 121 & 96 & 99 & 3.8 & 0.63M & \better{-54} & 0.44 & \better{-61} & 11.5 & \better{-26} & 10.5 & \better{-27} \\
\midrule

\multirow{3}{*}{\textbf{GPT-5.2}}
& \cellcolor{baseblue}Base
& 109 & 87 & -- & -- & 1.23M & -- & 1.77 & -- & 20.6 & -- & 19.4 & -- \\
& \cellcolor{skillgreen}Skill
& 114 & 90 & 84 & 3.8 & 0.26M & \better{-79} & 0.43 & \better{-75} & 9.9 & \better{-52} & 8.9 & \better{-54} \\
& \cellcolor{iteratepurple}Hier
& 100 & 79 & 95 & 3.6 & 0.60M & \better{-51} & 0.81 & \better{-48} & 11.8 & \better{-42} & 10.8 & \better{-44} \\
\bottomrule
\end{tabular}%
} 

\label{tab:three_mode}
\end{table}
\definecolor{LvlZero}{HTML}{AA7876} 
\definecolor{LvlOne}{HTML}{799C74}  
\definecolor{LvlTwo}{HTML}{587EB8}  
\definecolor{LvlThree}{HTML}{6F6C9C}

\newcommand{\lvlbox}[2]{%
  \begingroup
  \setlength{\fboxsep}{0.8pt}%
  \colorbox{#1}{\strut\,#2\,}%
  \endgroup
}

\begin{figure*}[t]
    \centering
    \includegraphics[width=\linewidth]{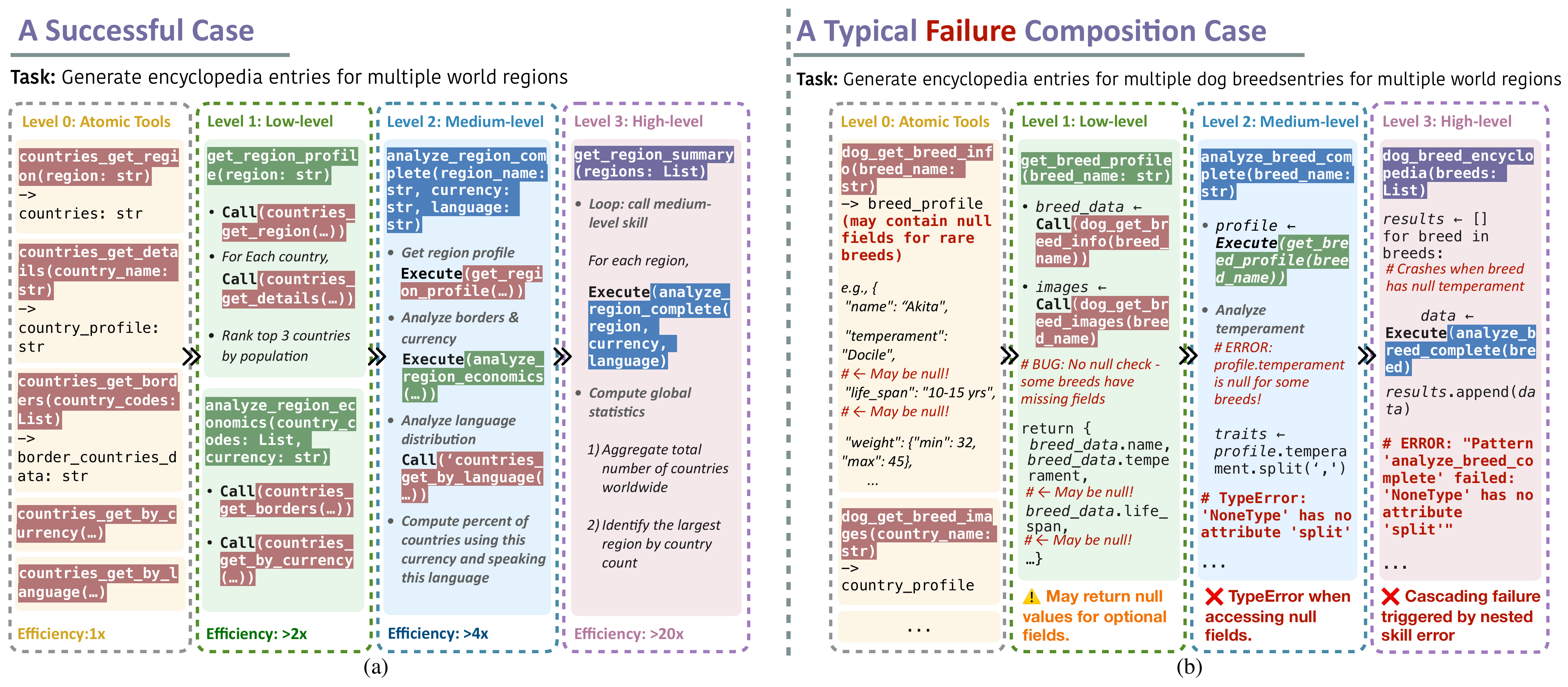}
    \vspace{-15pt}
    \caption{
    (a) Hierarchical skill composition in Iteration mode. A task organized as a depth-3 skill hierarchy, where
\protect\lvlbox{LvlZero}{\textcolor{white}{atomic tools}} are encapsulated by
\protect\lvlbox{LvlOne}{\textcolor{white}{low-level skills}}, composed into
\protect\lvlbox{LvlTwo}{\textcolor{white}{medium-level skills}}  with additional processing, and orchestrated by a
\protect\lvlbox{LvlThree}{\textcolor{white}{high-level skill}}. Efficiency gains compound across levels.
    (b) Error propagation in hierarchical skills. A null value returned by a low-level skill triggers a \texttt{TypeError} in the medium-level skill, which cascades into complete failure of the high-level skill. The tree structure amplifies the impact of edge-case bugs.}

    \label{fig:cases}
\vspace{-10pt}
\end{figure*}


\paragraph{Practical Challenges.} However, our experiments reveal that Hierarchical mode exhibits \textit{lower overall success rates} compared to flat Skill mode. The primary reason is \textbf{error propagation through the skill hierarchy}. Figure~\ref{fig:cases} (a) illustrates a typical failure pattern: a low-level skill (\texttt{get\_breed\_profile}) returns data with null fields for edge cases, which propagates upward and causes a \texttt{TypeError} in the medium-level skill (\texttt{analyze\_breed\_complete}), ultimately cascading into complete failure of the high-level skill (\texttt{compile\_breed\_encyclopedia}).

To illustrate cascading failures arising from implementation details, we identify three underlying micro-level factors: (1) compounding failures, where a skill at depth $d$ depends on its entire dependency subtree and success rate degrades rapidly with nesting; (2) latent bugs, where early-created skills may harbor edge-case errors that only manifest upon reuse, contaminating all higher-level skills built upon them; and (3) debugging overhead, where diagnosing nested failures requires tracing through dependencies—a cost that often exceeds simply re-executing with flat tool calls.

\paragraph{Empirical Results.} Table~\ref{tab:three_mode} compares Base, flat Skill, and hierarchical composition. Overall, deeper composition is \emph{not} a consistently beneficial scaling strategy. For a strong model (GPT-5.2), moving from flat Skill to Hierarchy reduces end-to-end success from 90\% to 79\%, while also weakening token savings (0.26M vs.\ 0.60M). Even when success does not change (e.g., Claude-4.5-Sonnet remains at 96\% in both modes), Hierarchy can still be less efficient (0.40M vs.\ 0.63M). Notably, Hierarchy often achieves high \textit{Exec} rates (e.g., 95--99\%), yet this does not translate into higher task success. Together, these results suggest that \textit{shallow, well-tested skill libraries} are currently more reliable and cost-effective than \textit{deep, automatically generated hierarchies}; realizing the latter likely requires much stronger systematic error handling and compositional verification.

\subsection{Cross-task Generalization}
A key property of a useful composition is its ability to generalize across problem difficulty. If a Skill captures reusable procedural structure rather than instance-specific solutions, it should transfer from simpler to more complex tasks (and vice versa) within the same task family. We therefore evaluate whether Skills learned at one difficulty level can be effectively reused at other difficulty levels.

We implement \textbf{Cross-Task Mode} using a two-phase static transfer approach. In Phase 1 (Skill Creation), an agent solves tasks at the \textit{source} difficulty level in standard Skill mode, creating and caching Skills in its workspace. In Phase 2 (Skill Transfer), the runner: (1) copies the pre-computed Skill cache to the workspace for \textit{target} difficulty tasks, (2) generates a \texttt{cross\_task\_skills\_summary} and injects into the system prompt, providing the agent with a structured description of available Skills including signatures, parameters, and execution history, and (3) executes the agent on target tasks with full access to the inherited Skills.

We evaluate three transfer directions: \textbf{Easy$\rightarrow$Hard} (Skills from e1--e3 tasks transferred to h1 tasks), \textbf{Hard$\rightarrow$Easy} (Skills from hard tasks applied to easy tasks), and \textbf{Hard$\rightarrow$Hard} (Skills from one set of hard tasks applied to different hard tasks within the same family). This static transfer approach isolates the generalization capability of Skills by preventing any modification or accumulation during Phase 2 execution.


\definecolor{baseblue}{RGB}{230, 240, 250}
\definecolor{skillgreen}{RGB}{230, 245, 235}

\begin{table}[t]
\centering
\Large
\renewcommand{\arraystretch}{1.5}
\setlength{\tabcolsep}{4pt}

\caption{Cross-task skill generalization results. \textbf{E$\rightarrow$H}: Skills learned from easy tasks (e1--e3) transferred to hard tasks (h1). \textbf{H$\rightarrow$E}: Skills from hard tasks applied to easy tasks. \textbf{H$\rightarrow$H}: Skills from hard tasks reapplied to the same hard tasks. \textbf{Base}: Baseline without skill transfer. \textbf{Skill}: With cross-task skill transfer. \textbf{Skill Exec}: Skill execution success rate. Efficiency metrics computed over tasks where both modes succeeded.~\textbf{Claude} is Claude-4.5-Sonnet and~\textbf{Gemini} is Gemini-3-Pro. Avg Tokens are in millions.
}

\resizebox{\columnwidth}{!}{%
\begin{tabular}{ll|cc|c|ccc|ccc}
\toprule
\multirow{2}{*}{\textbf{Model}} &
\multirow{2}{*}{\textbf{Setting}} &
\multicolumn{2}{c|}{\textbf{Success Rate}} &
\textbf{Skill} &
\multicolumn{3}{c|}{\textbf{Avg Tokens}} &
\multicolumn{3}{c}{\textbf{Avg Cost (\$)}} \\
\cmidrule{3-4}\cmidrule{6-8}\cmidrule{9-11}
& &
\cellcolor{baseblue}\textbf{Base} &
\cellcolor{skillgreen}\textbf{Skill} &
\textbf{Exec} &
\cellcolor{baseblue}\textbf{Base} &
\cellcolor{skillgreen}\textbf{Skill} &
\textbf{Diff} &
\cellcolor{baseblue}\textbf{Base} &
\cellcolor{skillgreen}\textbf{Skill} &
\textbf{Diff} \\
\midrule
\textbf{Claude} & E$\rightarrow$H & 20/21 (95\%) & 21/21 (100\%) & 100\% & 1.92 & 1.56 & \better{-19\%} & 1.41 & 1.07 & \better{-24\%} \\
 & H$\rightarrow$E & 60/63 (95\%) & 60/63 (95\%) & 97\% & 1.06 & 0.69 & \better{-35\%} & 0.81 & 0.44 & \better{-45\%} \\
 & H$\rightarrow$H & 20/21 (95\%) & 20/21 (95\%) & 98\% & 1.96 & 0.47 & \better{-76\%} & 1.46 & 0.43 & \better{-71\%} \\
\midrule
\textbf{Gemini} & E$\rightarrow$H & 16/21 (76\%) & 19/21 (90\%) & 99\% & 1.33 & 0.78 & \better{-41\%} & 1.26 & 0.76 & \better{-39\%} \\
 & H$\rightarrow$E & 55/63 (87\%) & 60/63 (95\%) & 100\% & 0.55 & 0.36 & \better{-35\%} & 0.46 & 0.30 & \better{-35\%} \\
 & H$\rightarrow$H & 16/21 (76\%) & 21/21 (100\%) & 99\% & 1.30 & 0.75 & \better{-42\%} & 1.23 & 0.67 & \better{-46\%} \\
\bottomrule
\end{tabular}%
} 
\label{tab:cross_task}
\end{table}
\vspace{-10pt}
\paragraph{Empirical Results.} Table~\ref{tab:cross_task} studies cross-task transfer across difficulty. For Claude-4.5-Sonnet, Easy$\rightarrow$Hard tasks raise success from 95\% to 100\% and cut tokens from 1.92M to 1.56M, and Hard$\rightarrow$Hard keeps success at 95\% while dropping tokens from 1.96M to 0.47M. For Gemini-3-Pro, transfer also improves both success and efficiency. Easy$\rightarrow$Hard increases success from 76\% to 90\% and reduces tokens from 1.33M to 0.78M. Hard$\rightarrow$Hard increases success from 76\% to 100\% and reduces tokens from 1.30M to 0.75M. Notably, transferred Skills execute with consistently high \textit{Exec} (typically 97--100\%), suggesting that a pre-computed Skill cache learned at one level can be reused across other levels with strong cross-task generalization.

\subsection{Cross-Model Skill Generalization}
\label{sec:cross-model}

\begin{figure}[h!]
    \centering
    \begin{subfigure}[b]{0.45\textwidth}
        \centering
        \includegraphics[width=\textwidth]{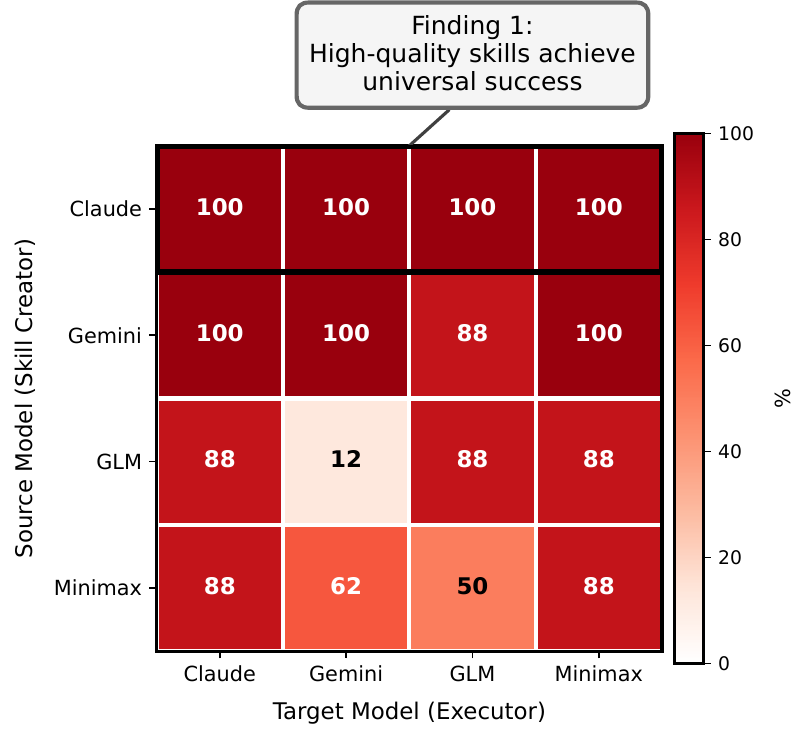}
        \caption{Success Rate}
        \label{fig:cross-model-success}
    \end{subfigure}
    \hfill
    \begin{subfigure}[b]{0.45\textwidth}
        \centering
        \includegraphics[width=\textwidth]{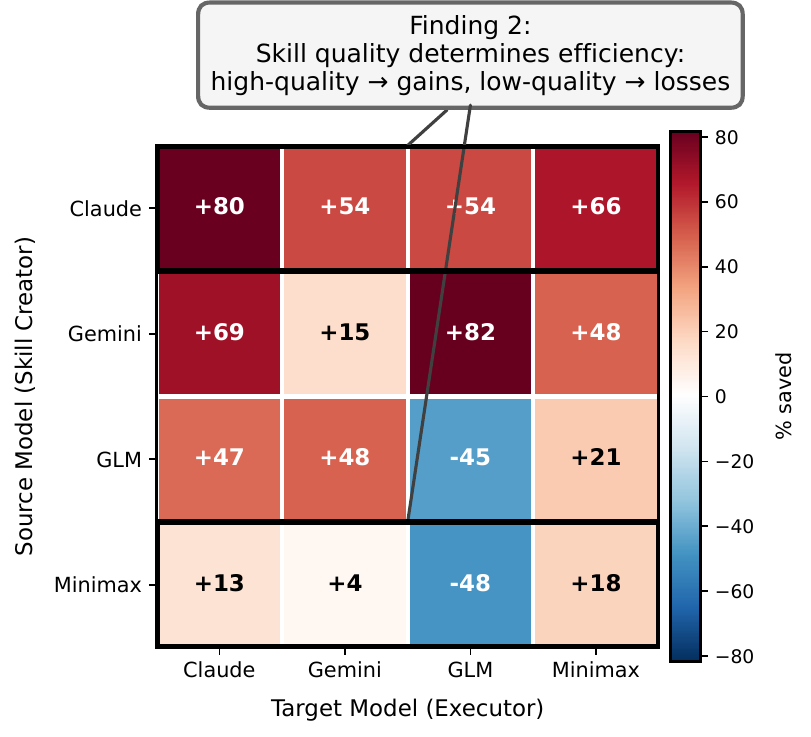}
        \caption{Token Saving}
        \label{fig:cross-model-token}
    \end{subfigure}
    \caption{Cross-model skill reuse heatmaps. Each cell $(i,j)$ shows the result when model $j$ executes skills created by model $i$. Bold borders highlight key findings. Token saving uses a diverging colormap: \textcolor{blue!70!black}{blue} = increased cost, \textcolor{red!70!black}{red} = reduced cost.}
    \label{fig:cross-model-heatmap}
\end{figure}

To investigate whether skills created by one model can benefit other models, we conduct a cross-model static reuse experiment on 8 hard-difficulty tasks. Four models (Claude, Gemini, GLM, and Minimax) each create skills during their initial task execution, and these skills are then provided to all four models for execution in static-reuse mode, resulting in a total of 16 cross-model combinations (including 4 self-reuse baselines). In static-reuse mode, agents can invoke pre-loaded skills via \texttt{execute\_skill} but cannot create new skills, ensuring that performance differences reflect skill quality and cross-model compatibility rather than on-the-fly adaptation. Figure~\ref{fig:cross-model-heatmap} presents the results as two heatmaps: task success rate and token saving percentage.

\paragraph{Finding 1: High-quality skills achieve universal success.}
The first row of Figure~\ref{fig:cross-model-success} demonstrates that Claude-created skills achieve 100\% success rate across all four target models, including when executed by Gemini, GLM, and Minimax. This universally high success rate indicates that well-abstracted skills with clear parameter interfaces transfer effectively across different executor models, regardless of their architectural differences.

\paragraph{Finding 2: Skill quality determines efficiency gain or loss.}
Figure~\ref{fig:cross-model-token} reveals a stark contrast in computational efficiency based on skill creator quality. Claude-created skills (first row) yield consistently high token savings of 54--81\% across all executors, demonstrating that high-quality skills provide universal efficiency benefits. In contrast, Minimax-created skills (bottom row) result in token savings ranging from $-48\%$ to $+18\%$, meaning poorly designed skills often \emph{increase} rather than decrease computational cost. Notably, self-reuse (diagonal) does not always outperform cross-model reuse: Claude achieves 69.2\% saving with Gemini's skills, substantially exceeding Gemini's own 14.8\% self-reuse---indicating that executor capability can compensate for moderate skill quality, but cannot salvage fundamentally flawed skill designs.

\paragraph{Implications.}
These findings demonstrate that \emph{skill creator quality matters more than executor capability}: investing in high-quality skill creation from capable models yields transferable efficiency benefits across the entire model ecosystem, while poorly designed skills can harm performance regardless of which model executes them. This suggests that multi-agent systems should prioritize skill libraries curated from high-capability models rather than allowing arbitrary skill contributions from all participants.

\vspace{10pt}
\section{Conclusion}
We introduced~\textbf{SkillCraft}, a benchmark containing 126 tasks with recurring substructures, and Skill Mode, a protocol enabling agents to auto compose, cache, and reuse tool sequences. This framework allows us to measure whether agents can acquire compositional skills rather than merely execute isolated tool calls. Evaluating state-of-the-art models reveals two key findings. First, skill reuse reduces token usage by up to 80\% while maintaining or improving success rates, Second, Efficiency gains strongly correlated to model intelligence. Besides, skills generalize well across tasks and models ($>$95\% execution), though hierarchical compositions are less reliable due to error accumulation across nested steps, highlighting compositional skill acquisition as a crucial capability for robust long-horizon tool use.

\section*{Impact Statement}

This paper presents work whose goal is to advance the field of machine learning. There are many potential societal consequences of our work, none of which we feel must be specifically highlighted here.

\bibliography{example_paper}
\bibliographystyle{icml2026}

\newpage
\appendix
\appendix


\section{Related Work}

Tool-use benchmarks mainly differ in the realism of tool executability and in whether tasks require long-horizon composition. In controlled settings, BFCL~\citep{patilberkeley} reduces tool use to structured function-parameter prediction, while $\tau$-Bench and ACEBench emphasize multi-turn interaction and correct tool selection under reproducible environments~\citep{yao2024tau, chen2025acebench}. Gorilla and AgentBench broaden tool and domain coverage~\citep{patil2023gorilla, liu2023agentbench}, but primarily evaluate API selection, such that short tool-call chains often suffice.

More realistic benchmarks execute tools in richer environments. AppWorld supports application-level state transitions~\citep{trivedi2024appworld}, and MCP-based suites such as MCPWorld, MCP-RADAR, MCPEval, MCP-AgentBench, LiveMCPBench, and MCPAtlas standardize tool integration across servers~\citep{yan2025mcpworld, gao2025mcp, liu2025mcpeval, guo2025mcp, mo2025livemcpbench, bandimcp}, though tasks often remain single-application with simplified initial states. WebArena, OSWorld, SWE-Bench, and TheAgentCompany emphasize long-horizon execution and error recovery in web, desktop, and code workflows~\citep{zhou2023webarena, xie2024osworld, jimenez2024swebench, xu2024theagentcompanybenchmarkingllmagents}, while GAIA, ARE, and BrowseComp focus on everyday tasks and web-based information seeking~\citep{mialon2023gaia, froger2025scaling, wei2025browsecomp}. Tool Decathlon (Toolathlon) further consolidates real tools, fuzzy instructions, execution verification, and cross-application workflows~\citep{li2025tool}.

On the pipeline side, most tool-using agents follow the “reasoning–acting–observing" loop introduced by ReAct~\citep{yao2023react}, where planning and state tracking are repeated at every tool call. CodeAct~\citep{wang2024executable} shifts the action space to executable code to express control flow and multi-tool orchestration, but it still regenerates code per task and does not accumulate reusable procedures. Voyager~\citep{wang2023voyager} and Ghost in the Minecraft~\citep{zhu2023ghost} show that agents can grow a code skill library through exploration, yet the resulting skills are tied to game rules and state spaces. CREATOR~\citep{qian2023creator} abstracts reusable components from patterns but provides limited evidence of robust cross-task generalization in realistic tool ecosystems. 
Anthropic Skills~\citep{anthropic_agentskills} packages workflows as explicit skill modules, but these modules are typically authored and configured by humans rather than induced from execution. In contrast, SkillCraft enables autonomous reuse with a minimal MCP protocol that compiles successful tool sequences into verified executable skills.

\section{Skill Mode: System Details}
\label{app:system}

\subsection{Four primitive tools enabling Skill Mode}
\label{app:primitive_tools}
\begin{figure*}[t!]
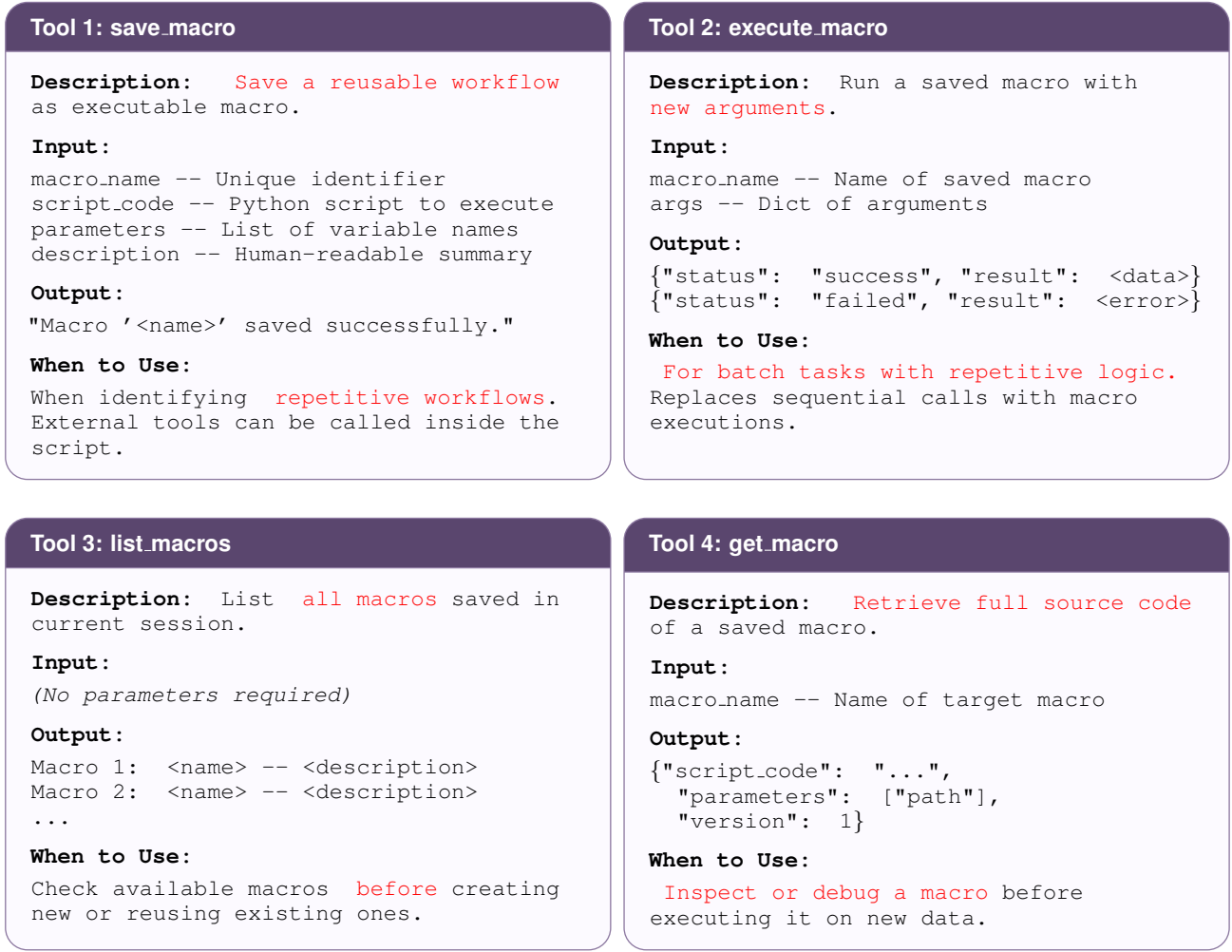

\centering
\setlength{\tabcolsep}{0pt}
\begin{tabular}{@{}p{0.495\textwidth}@{\hspace{0.01\textwidth}}p{0.495\textwidth}@{}}

\begin{toolbox}[equal height group=row1]{Tool 1: save\_macro}\footnotesize\color{black}
\textbf{Description:} ~\textcolor{red}{Save a reusable workflow} as executable macro.

\vspace{2mm}
\textbf{Input:}\\[1mm]
\small
\texttt{macro\_name} -- Unique identifier\\
\texttt{script\_code} -- Python script to execute\\
\texttt{parameters} -- List of variable names\\
\texttt{description} -- Human-readable summary

\vspace{2mm}
\textbf{Output:}\\[1mm]
\small\texttt{"Macro '<name>' saved successfully."}

\vspace{2mm}
\textbf{When to Use:}\\[1mm]
\small When identifying ~\textcolor{red}{repetitive workflows}. External tools can be called inside the script.
\end{toolbox}
&
\begin{toolbox}[equal height group=row1]{Tool 2: execute\_macro}\footnotesize\color{black}
\textbf{Description:} Run a saved macro with ~\textcolor{red}{new arguments}.

\vspace{2mm}
\textbf{Input:}\\[1mm]
\small
\texttt{macro\_name} -- Name of saved macro\\
\texttt{args} -- Dict of arguments

\vspace{2mm}
\textbf{Output:}\\[1mm]
\small
\texttt{\{"status": "success", "result": <data>\}}\\
\texttt{\{"status": "failed", "result": <error>\}}

\vspace{2mm}
\textbf{When to Use:}\\[1mm]
\small ~\textcolor{red}{For batch tasks with repetitive logic.} Replaces sequential calls with macro executions.
\end{toolbox}
\\[4mm]

\begin{toolbox}[equal height group=row2]{Tool 3: list\_macros}\footnotesize\color{black}
\textbf{Description:} List ~\textcolor{red}{all macros} saved in current session.

\vspace{2mm}
\textbf{Input:}\\[1mm]
\small\textit{(No parameters required)}

\vspace{2mm}
\textbf{Output:}\\[1mm]
\small
\texttt{Macro 1: <name> -- <description>}\\
\texttt{Macro 2: <name> -- <description>}\\
\texttt{...}

\vspace{2mm}
\textbf{When to Use:}\\[1mm]
\small Check available macros ~\textcolor{red}{before} creating new or reusing existing ones.
\end{toolbox}
&
\begin{toolbox}[equal height group=row2]{Tool 4: get\_macro}\footnotesize\color{black}
\textbf{Description:} ~\textcolor{red}{Retrieve full source code} of a saved macro.

\vspace{2mm}
\textbf{Input:}\\[1mm]
\small\texttt{macro\_name} -- Name of target macro

\vspace{2mm}
\textbf{Output:}\\[1mm]
\small
\texttt{\{"script\_code": "...",}\\
\texttt{~~"parameters": ["path"],}\\
\texttt{~~"version": 1\}}

\vspace{2mm}
\textbf{When to Use:}\\[1mm]
\small ~\textcolor{red}{Inspect or debug a macro} before executing it on new data.
\end{toolbox}
\\
\end{tabular}
\caption{Pseudo-code for the four primitive tools that enable Skill Mode.}
\label{fig:macro-tools}
\end{figure*}

We illustrate the detailed design and functionality of the four primitive tools that together enable the proposed Skill Mode in Figure~\ref{fig:macro-tools}.

\subsection{Why Skill Mode improves efficiency}
\label{app:why_efficiency}
\begin{figure*}[t!]
    \centering
    \includegraphics[width=\textwidth]{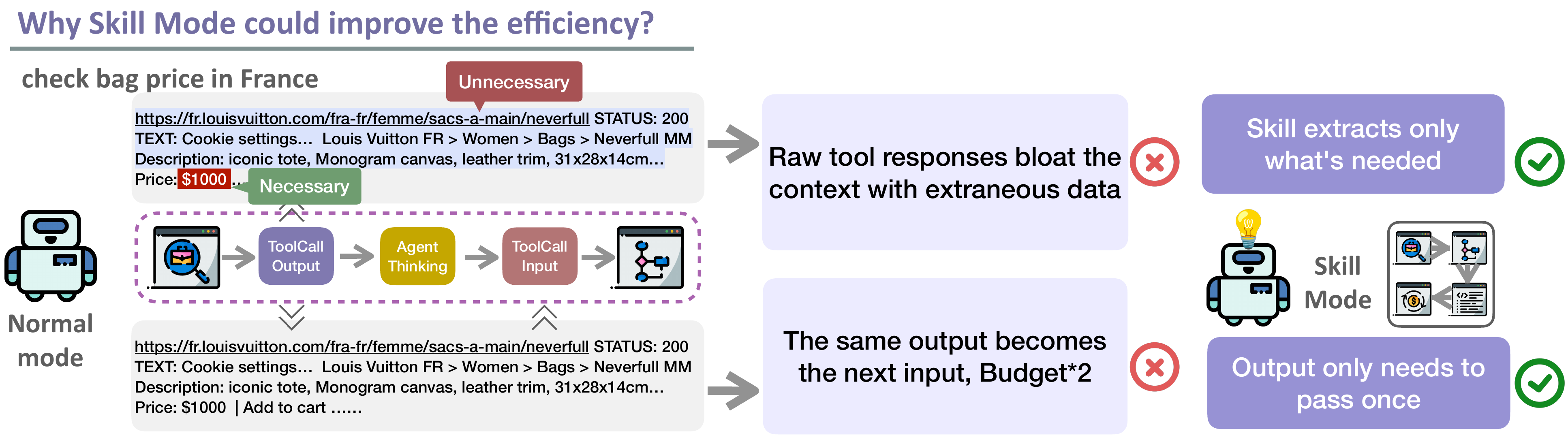}
    \caption{Skill Mode improves efficiency through two mechanisms. First, it reduces argument passing overhead by enabling direct tool chaining (Tool A $\rightarrow$ Tool B $\rightarrow$ Tool C) rather than shuttling intermediate outputs through the agent (Tool A $\rightarrow$ Agent $\rightarrow$ Tool B $\rightarrow$ Agent $\rightarrow$ Tool C). Second, it amortizes planning cost by allowing agents to reuse previously discovered tool sequences, eliminating the need to reason about recurring multi-step patterns from scratch.}
    \label{fig:why-figure}
\end{figure*}

Figure~\ref{fig:why-figure} illustrates why Skill Mode improves efficiency through two complementary mechanisms. In normal tool use, raw tool outputs (e.g., full webpages or verbose API responses) are repeatedly injected into the context, bloating the prompt with extraneous information and incurring repeated argument-passing costs as the output of one tool becomes the input of the next via the agent. Skill Mode instead extracts and caches only the minimal, task-relevant fields, enabling direct tool-to-tool chaining and allowing intermediate results to be passed once rather than re-serialized at every step. Moreover, by reusing previously discovered tool sequences as atomic skills, the agent amortizes planning and reasoning cost over repeated executions, avoiding the need to reconstruct the same multi-step workflow from scratch.

\subsection{Implementation details}
\label{app:implementation_details}
This section provides additional implementation details that complement the methodology described in the main text.

\paragraph{Execution Configuration.}
To ensure reproducibility and prevent resource exhaustion, we impose several execution limits on each task. Each task is allocated a maximum of 150 conversation turns (or 300 steps in single-turn mode) and a 60-minute timeout. We enforce cumulative token limits of 1M input tokens and 150K output tokens per task, with individual requests capped at 150K input tokens. Tasks exceeding these limits are terminated and evaluated based on partial completion. For generation, all models use temperature$=$0.0 and top\_p$=$1.0 to ensure deterministic outputs. We set \texttt{tool\_choice="auto"} to allow models to decide when to invoke tools autonomously.

\paragraph{Skill Storage and Execution.}
Skills are persisted as JSON entries in a \texttt{skill\_cache.json} file within each task's workspace. Each skill entry contains: (1) \texttt{script\_code}---executable Python code that invokes tools via a \texttt{call\_tool(name, **kwargs)} interface, (2) \texttt{parameters}---a list of input parameter names, (3) \texttt{description}---natural language documentation, and (4) \texttt{execution\_stats}---runtime statistics tracking successful and failed executions.

\paragraph{Evaluation Protocol.}
We employ a partial-credit scoring system where each task defines multiple weighted evaluation criteria. Typical criteria include: output file existence (10 points), JSON validity (10 points), data completeness (30 points), and field-level accuracy (50 points). A task is considered \emph{successful} if it achieves $\geq$90\% of the maximum score. Efficiency metrics (tokens, cost, turns, tool calls) are computed only over tasks where \emph{both} baseline and skill modes succeed, ensuring fair comparison. All API costs are tracked via the OpenRouter billing API.

\section{SkillCraft: Benchmark Construction Details}
\label{app:benchmark}

\subsection{Task API Sources}
\label{app:task_sources}

\definecolor{urlblue}{RGB}{0, 180, 200}
\begin{table*}[h]
\centering
\small
\renewcommand{\arraystretch}{1.15}
\setlength{\tabcolsep}{5pt}

\caption{Complete list of API sources used in \textsc{SkillCraft}. The benchmark comprises 21 task families across 6 domains (Entertainment, Reference, Education, Developer, Science, Food). \textbf{Tools}: number of distinct API-wrapping functions per task. Each task family includes 6 difficulty-scaled variants (Easy: 3 subtasks, Medium: 5 subtasks, Hard: 7 subtasks), totaling 126 tasks. All APIs except Local DNA Analysis are publicly available REST endpoints.}

\begin{tabular}{llcl}
\toprule
\textbf{Task Family} & \textbf{Domain} & \textbf{Tools} & \textbf{Source} \\
\midrule
Cat Facts Collector & Reference & 5 & \textcolor{urlblue}{\url{https://catfact.ninja}} \\
Cocktail Menu Generator & Food & 5 & \textcolor{urlblue}{\url{https://thecocktaildb.com}} \\
Countries Encyclopedia & Reference & 5 & \textcolor{urlblue}{\url{https://restcountries.com}} \\
D\&D Campaign Builder & Gaming & 6 & \textcolor{urlblue}{\url{https://dnd5eapi.co}} \\
D\&D Monster Compendium & Gaming & 6 & \textcolor{urlblue}{\url{https://dnd5eapi.co}} \\
Dog Breeds Encyclopedia & Reference & 5 & \textcolor{urlblue}{\url{https://dog.ceo/api}} \\
GitLab Deep Analysis & Developer & 6 & \textcolor{urlblue}{\url{https://gitlab.com/api/v4}} \\
Jikan Anime Analysis & Entertainment & 5 & \textcolor{urlblue}{\url{https://api.jikan.moe}} \\
JSONPlaceholder Analyzer & Developer & 7 & \textcolor{urlblue}{\url{https://jsonplaceholder.typicode.com}} \\
Local DNA Analysis & Science & 5 & Custom Implementation \\
Name Demographics & Society & 5 & \textcolor{urlblue}{\url{https://genderize.io}} \\
Open-Meteo Weather & Science & 5 & \textcolor{urlblue}{\url{https://open-meteo.com}} \\
PokéAPI Pokédex & Gaming & 5 & \textcolor{urlblue}{\url{https://pokeapi.co}} \\
Random User Database & Society & 5 & \textcolor{urlblue}{\url{https://randomuser.me}} \\
Recipe Cookbook Builder & Food & 6 & \textcolor{urlblue}{\url{https://themealdb.com}} \\
Rick \& Morty Explorer & Entertainment & 5 & \textcolor{urlblue}{\url{https://rickandmortyapi.com}} \\
TVMaze Series Analyzer & Developer & 5 & \textcolor{urlblue}{\url{https://api.tvmaze.com}} \\
University Directory & Education & 5 & \textcolor{urlblue}{\url{http://universities.hipolabs.com}} \\
USGS Earthquake Monitor & Science & 6 & \textcolor{urlblue}{\url{https://earthquake.usgs.gov}} \\
Vocabulary Builder & Reference & 5 & \textcolor{urlblue}{\url{https://dictionaryapi.dev}} \\
World Bank Snapshot & Education & 5 & \textcolor{urlblue}{\url{https://api.worldbank.org}} \\
\bottomrule
\end{tabular}

\label{tab:task_sources}
\end{table*}

We present the complete list of API sources used in \textsc{SkillCraft} benchmark in Table~\ref{tab:task_sources}. 
Our 21 task families span six application domains—from entertainment and gaming to science and development—covering a diverse range of real-world API interaction patterns.
All APIs are publicly available REST endpoints that require structured multi-step interactions, making them ideal candidates for evaluating skill composition and reuse.
For each task family, we implement 5--7 tool functions wrapping distinct API endpoints; difficulty levels (Easy/Medium/Hard) control the number of subtasks (3/4/5) and thus total API calls required per task.
Most of these APIs are sourced from existing community-maintained projects, while the Local DNA Analysis task uses a custom implementation for bioinformatics operations.

\section{Additional Analyses}
\label{app:analysis}

\subsection{Results by task difficulty}
\label{app:difficulty_analysis}
\begin{table*}[h]
\centering
\small
\renewcommand{\arraystretch}{1.15}
\setlength{\tabcolsep}{3pt}
\caption{
Results breakdown by difficulty level (Easy: e1--e3, Medium: m1--m2, Hard: h1).
\textbf{Success Rate}: task completion rate (score $\geq$ 90) for \colorbox{baseblue}{Baseline} and \colorbox{skillgreen}{Skill} modes.
\textbf{Skill Stats}: Exec = skill execution success rate; Reuse = average times each skill is invoked.
\textbf{Efficiency metrics}: per-task averages computed over tasks where \emph{both} modes succeeded.
\textbf{Diff}: percentage change; \better{negative} = improvement, \worse{positive} = degradation.
}

\resizebox{\textwidth}{!}{%
\begin{tabular}{l|l|cc|cc|ccc|ccc|ccc|ccc}
\toprule
\multirow{2}{*}{\textbf{Model}} &
\multirow{2}{*}{\textbf{Diff.}} &
\multicolumn{2}{c|}{\textbf{Success Rate}} &
\multicolumn{2}{c|}{\textbf{Skill Stats}} &
\multicolumn{3}{c|}{\textbf{Avg Tokens}} &
\multicolumn{3}{c|}{\textbf{Avg Cost (\$)}} &
\multicolumn{3}{c|}{\textbf{Avg Turns}} &
\multicolumn{3}{c}{\textbf{Avg Tool Calls}} \\
\cmidrule{3-4}\cmidrule{5-6}\cmidrule{7-9}\cmidrule{10-12}\cmidrule{13-15}\cmidrule{16-18}
& &
\cellcolor{baseblue}\textbf{Base} & \cellcolor{skillgreen}\textbf{Skill} &
\textbf{Exec} & \textbf{Reuse} &
\cellcolor{baseblue}\textbf{Base} & \cellcolor{skillgreen}\textbf{Skill} & \textbf{Diff} &
\cellcolor{baseblue}\textbf{Base} & \cellcolor{skillgreen}\textbf{Skill} & \textbf{Diff} &
\cellcolor{baseblue}\textbf{Base} & \cellcolor{skillgreen}\textbf{Skill} & \textbf{Diff} &
\cellcolor{baseblue}\textbf{Base} & \cellcolor{skillgreen}\textbf{Skill} & \textbf{Diff} \\
\midrule
\multirow{3}{*}{\textbf{Kimi-K2-Thinking}} & Easy & 30/63 (48\%) & 32/63 (51\%) & 81\% & 2.6$\times$ & 427K & 293K & \better{-31\%} & 0.17 & 0.12 & \better{-29\%} & 9.0 & 13.7 & \worse{+53\%} & 12.2 & 10.6 & \better{-13\%} \\
 & Medium & 17/42 (40\%) & 17/42 (40\%) & 76\% & 2.9$\times$ & 576K & 335K & \better{-42\%} & 0.23 & 0.14 & \better{-39\%} & 8.1 & 15.1 & \worse{+87\%} & 19.5 & 12.5 & \better{-36\%} \\
 & Hard & 8/21 (38\%) & 7/21 (33\%) & 66\% & 4.8$\times$ & 622K & 285K & \better{-54\%} & 0.25 & 0.13 & \better{-50\%} & 10.4 & 16.4 & \worse{+58\%} & 27.8 & 18.4 & \better{-34\%} \\
\midrule
\multirow{3}{*}{\textbf{DeepSeek-V3.2-EXP}} & Easy & 42/63 (66\%) & 47/63 (74\%) & 94\% & 2.5$\times$ & 943K & 512K & \better{-46\%} & 0.27 & 0.22 & \better{-18\%} & 28.2 & 25.7 & \better{-9\%} & 15.4 & 13.2 & \better{-14\%} \\
 & Medium & 25/42 (59\%) & 25/42 (59\%) & 89\% & 3.2$\times$ & 1.34M & 556K & \better{-59\%} & 0.22 & 0.08 & \better{-63\%} & 36.2 & 30.4 & \better{-16\%} & 23.4 & 17.2 & \better{-26\%} \\
 & Hard & 9/21 (42\%) & 15/21 (71\%) & 88\% & 4.4$\times$ & 844K & 547K & \better{-35\%} & 0.22 & 0.07 & \better{-69\%} & 21.4 & 33.1 & \worse{+55\%} & 28.7 & 17.6 & \better{-39\%} \\
\midrule
\multirow{3}{*}{\textbf{DeepSeek-R1}} & Easy & 50/63 (79\%) & 55/63 (87\%) & 68\% & 2.9$\times$ & 498K & 470K & \better{-6\%} & 0.21 & 0.20 & \better{-3\%} & 14.0 & 16.8 & \worse{+20\%} & 11.1 & 11.0 & \better{-1\%} \\
 & Medium & 28/42 (66\%) & 31/42 (74\%) & 77\% & 3.5$\times$ & 631K & 241K & \better{-62\%} & 0.26 & 0.13 & \better{-50\%} & 13.6 & 14.9 & \worse{+10\%} & 15.8 & 10.7 & \better{-32\%} \\
 & Hard & 11/21 (52\%) & 15/21 (71\%) & 68\% & 4.2$\times$ & 855K & 421K & \better{-51\%} & 0.34 & 0.17 & \better{-49\%} & 17.1 & 16.9 & \better{-1\%} & 16.7 & 16.8 & \worse{+1\%} \\
\midrule
\multirow{3}{*}{\textbf{GLM-4.7}} & Easy & 52/63 (82\%) & 57/63 (90\%) & 90\% & 2.9$\times$ & 661K & 428K & \better{-35\%} & 0.18 & 0.12 & \better{-33\%} & 12.2 & 11.4 & \better{-6\%} & 13.6 & 11.8 & \better{-13\%} \\
 & Medium & 27/42 (64\%) & 36/42 (85\%) & 89\% & 4.0$\times$ & 874K & 514K & \better{-41\%} & 0.22 & 0.10 & \better{-54\%} & 13.7 & 15.3 & \worse{+11\%} & 19.0 & 15.1 & \better{-21\%} \\
 & Hard & 12/21 (57\%) & 15/21 (71\%) & 94\% & 4.9$\times$ & 1.17M & 648K & \better{-45\%} & 0.28 & 0.16 & \better{-43\%} & 19.6 & 15.1 & \better{-23\%} & 28.2 & 16.5 & \better{-41\%} \\
\midrule
\multirow{3}{*}{\textbf{Gemini 3 Pro}} & Easy & 55/63 (87\%) & 59/63 (93\%) & 95\% & 2.3$\times$ & 534K & 300K & \better{-44\%} & 0.46 & 0.30 & \better{-35\%} & 15.4 & 19.6 & \worse{+27\%} & 12.8 & 9.2 & \better{-29\%} \\
 & Medium & 37/42 (88\%) & 40/42 (95\%) & 92\% & 2.7$\times$ & 730K & 323K & \better{-56\%} & 0.68 & 0.31 & \better{-55\%} & 16.6 & 18.4 & \worse{+11\%} & 19.5 & 10.6 & \better{-46\%} \\
 & Hard & 16/21 (76\%) & 17/19 (89\%) & 96\% & 3.3$\times$ & 970K & 227K & \better{-77\%} & 0.90 & 0.27 & \better{-70\%} & 22.2 & 20.3 & \better{-8\%} & 28.9 & 8.7 & \better{-70\%} \\
\midrule
\multirow{3}{*}{\textbf{Minimax-M2.1}} & Easy & 59/63 (94\%) & 58/63 (96\%) & 100\% & 3.0$\times$ & 379K & 363K & \better{-4\%} & 0.04 & 0.03 & \better{-13\%} & 7.7 & 7.4 & \better{-4\%} & 12.1 & 11.4 & \better{-5\%} \\
 & Medium & 40/42 (95\%) & 41/42 (98\%) & 84\% & 3.6$\times$ & 468K & 380K & \better{-19\%} & 0.05 & 0.05 & \better{-8\%} & 7.9 & 7.1 & \better{-11\%} & 18.6 & 16.8 & \better{-10\%} \\
 & Hard & 18/21 (86\%) & 20/21 (95\%) & 100\% & 3.0$\times$ & 479K & 409K & \better{-15\%} & 0.06 & 0.05 & \better{-4\%} & 8.2 & 7.2 & \better{-12\%} & 26.8 & 24.8 & \better{-8\%} \\
\midrule
\multirow{3}{*}{\textbf{Claude 4.5 Sonnet}} & Easy & 60/63 (95\%) & 60/63 (95\%) & 99\% & 3.0$\times$ & 1.06M & 399K & \better{-62\%} & 0.81 & 0.25 & \better{-69\%} & 19.2 & 17.4 & \better{-9\%} & 11.4 & 9.0 & \better{-21\%} \\
 & Medium & 39/42 (92\%) & 41/42 (98\%) & 100\% & 3.7$\times$ & 1.54M & 369K & \better{-76\%} & 1.32 & 0.27 & \better{-80\%} & 22.0 & 17.7 & \better{-19\%} & 15.7 & 9.0 & \better{-43\%} \\
 & Hard & 20/21 (95\%) & 20/21 (95\%) & 98\% & 4.7$\times$ & 1.96M & 440K & \better{-77\%} & 1.46 & 0.40 & \better{-72\%} & 25.5 & 19.9 & \better{-22\%} & 20.3 & 10.2 & \better{-50\%} \\
\midrule
\multirow{3}{*}{\textbf{GPT-5.2}} & Easy & 59/63 (94\%) & 60/63 (95\%) & 91\% & 3.0$\times$ & 939K & 196K & \better{-79\%} & 1.38 & 0.30 & \better{-78\%} & 22.3 & 15.1 & \better{-32\%} & 12.5 & 7.7 & \better{-38\%} \\
 & Medium & 34/42 (81\%) & 37/42 (88\%) & 95\% & 4.2$\times$ & 1.44M & 314K & \better{-78\%} & 2.10 & 0.48 & \better{-77\%} & 26.9 & 17.0 & \better{-37\%} & 21.0 & 8.9 & \better{-58\%} \\
 & Hard & 16/21 (76\%) & 17/21 (80\%) & 90\% & 4.3$\times$ & 1.86M & 405K & \better{-78\%} & 2.72 & 0.61 & \better{-77\%} & 31.4 & 20.5 & \better{-35\%} & 41.3 & 13.4 & \better{-68\%} \\
\bottomrule
\end{tabular}
}

\label{tab:difficulty_breakdown}
\end{table*}

Table~\ref{tab:difficulty_breakdown} presents a detailed breakdown of our experimental results across three difficulty levels: Easy (tasks e1--e3), Medium (tasks m1--m2), and Hard (task h1). We identify several noteworthy patterns that provide deeper insights into the behavior and benefits of skill reuse.

\paragraph{Skill Reuse Frequency Increases with Task Complexity.}
Across all models, the average skill reuse count shows a consistent upward trend with task difficulty. For Easy tasks, skills are invoked 2.3--3.0$\times$ on average, while Hard tasks see 3.0--4.9$\times$ reuse. This pattern reflects the compositional nature of our benchmark: harder tasks require more repeated API compositions, which naturally leads to more opportunities for skill reuse. Notably, GLM-4.7 achieves the highest reuse rate (4.9$\times$) on Hard tasks, demonstrating effective skill generalization across complex scenarios.

\paragraph{Efficiency Gains are More Pronounced on Harder Tasks.}
Token savings exhibit a clear correlation with task difficulty. For frontier models like Claude 4.5 Sonnet and GPT-5.2, token reduction on Hard tasks reaches 77--78\%, compared to 62--79\% on Easy tasks. Similarly, tool call reduction is most dramatic on Hard tasks: Gemini 3 Pro achieves a 70\% reduction on Hard versus 29\% on Easy, while GPT-5.2 shows 68\% versus 38\%. This suggests that skill reuse provides greater benefits when tasks involve more complex, multi-step API orchestrations---precisely the scenarios where manual tool composition becomes most costly.

\paragraph{Success Rate Improvements Favor Challenging Tasks.}
For models with moderate baseline performance, skill reuse disproportionately improves success rates on Hard tasks. DeepSeek-V3.2-EXP shows a remarkable +29 percentage point improvement on Hard tasks (from 42\% to 71\%) compared to only +8 points on Easy tasks. Similarly, DeepSeek-R1 improves by +19 points on Hard versus +7 points on Easy. This indicates that skills learned from easier variants effectively transfer to help models overcome challenges they would otherwise fail, validating the cross-difficulty generalization capability of our skill framework.

\paragraph{High-Capacity Models Benefit from Efficiency, Not Accuracy.}
Frontier models (Claude, GPT-5.2) already achieve $>$95\% success rates on Easy tasks in baseline mode, leaving little room for accuracy improvement. However, they show the largest efficiency gains: Claude achieves 72\% average token reduction, and GPT-5.2 achieves 78\%. In contrast, Minimax-M2.1, which exhibits highly efficient baseline behavior (only 379K--479K tokens per task), shows modest 4--19\% token savings. This suggests that skill reuse is most valuable for models whose baseline execution involves verbose, sequential API interactions.

\paragraph{Skill Execution Remains Robust Across Difficulties.}
Skill execution success rates remain consistently high (66--100\%) across all difficulty levels for most models, indicating that skills created during easier tasks transfer reliably to harder contexts. The lowest execution rates appear in Kimi-K2-Thinking (66\% on Hard) and DeepSeek-R1 (68\% on Easy/Hard), both of which employ extended reasoning that may occasionally conflict with deterministic skill execution patterns.

\subsection{Direct execution mode}
\label{app:direct_exec}
\definecolor{baseblue}{RGB}{230, 240, 250}
\definecolor{skillgreen}{RGB}{230, 245, 235}
\definecolor{directexecpurple}{RGB}{240, 230, 250}

\begin{table*}[!h]
\centering
\small
\renewcommand{\arraystretch}{1.15}
\setlength{\tabcolsep}{3.5pt}
\caption{Three-mode comparison (Base, Skill, Direct Exec) on 48-task subset. \textbf{Base}: No skill library. \textbf{Skill}: With skill library from previous runs. \textbf{Direct Exec}: Skills are directly executed without agent intervention. Efficiency metrics are computed over tasks where both Base and the respective mode succeeded.}
\resizebox{\textwidth}{!}{%
\begin{tabular}{l|l|cc|cc|cc|cc|cc|cc}
\toprule
\multirow{2}{*}{\textbf{Model}} & \multirow{2}{*}{\textbf{Mode}} & \multicolumn{2}{c|}{\textbf{Success Rate}} & \multicolumn{2}{c|}{\textbf{Skill Stats}} & \multicolumn{2}{c|}{\textbf{Avg Tokens}} & \multicolumn{2}{c|}{\textbf{Avg Cost (\$)}} & \multicolumn{2}{c|}{\textbf{Avg Turns}} & \multicolumn{2}{c}{\textbf{Avg Tool Calls}} \\
\cmidrule{3-4}\cmidrule{5-6}\cmidrule{7-8}\cmidrule{9-10}\cmidrule{11-12}\cmidrule{13-14}
 &  & \textbf{Succ} & \textbf{Rate} & \textbf{Exec} & \textbf{Reuse} & \textbf{Val} & \textbf{Diff} & \textbf{Val} & \textbf{Diff} & \textbf{Val} & \textbf{Diff} & \textbf{Val} & \textbf{Diff} \\
\midrule
\textbf{Claude-4.5-Sonnet} & \cellcolor{baseblue}Base & 47/48 & 98\% & -- & -- & 1.72M & -- & 1.73 & -- & 15.7 & -- & 14.7 & -- \\
 & \cellcolor{skillgreen}Skill & 43/48 & 90\% & 99\% & 3.7$\times$ & 0.34M & \better{-80\%} & 0.22 & \better{-87\%} & 10.5 & \better{-33\%} & 9.5 & \better{-36\%} \\
 & \cellcolor{directexecpurple}Direct Exec & 46/48 & 96\% & 68\% & 3.1$\times$ & 0.16M & \better{-90\%} & 0.17 & \better{-89\%} & 5.8 & \better{-64\%} & 4.8 & \better{-68\%} \\
\midrule
\textbf{GPT-5.2} & \cellcolor{baseblue}Base & 45/48 & 94\% & -- & -- & 1.18M & -- & 1.52 & -- & 24.5 & -- & 23.1 & -- \\
 & \cellcolor{skillgreen}Skill & 43/48 & 90\% & 97\% & 3.5$\times$ & 0.26M & \better{-78\%} & 0.39 & \better{-74\%} & 8.9 & \better{-64\%} & 7.9 & \better{-66\%} \\
 & \cellcolor{directexecpurple}Direct Exec & 41/48 & 85\% & 68\% & 3.1$\times$ & 0.06M & \better{-95\%} & 0.14 & \better{-91\%} & 4.5 & \better{-78\%} & 3.5 & \better{-81\%} \\
\bottomrule
\end{tabular}
}
\label{tab:direct_exec}
\end{table*}

We further investigate the efficiency impact of script parameterization by implementing \textbf{Direct Exec Mode}, an alternative approach that trades generalization capability for execution efficiency.

In our Skill mode, agents create parameterized skills through a two-step process: first \texttt{save\_skill} to store a reusable script with parameter placeholders, then \texttt{execute\_skill} to invoke it with specific arguments. This design enables skill reuse across similar tasks but introduces overhead from parameter abstraction and the save-then-execute workflow.

Direct Exec Mode takes a fundamentally different approach. Instead of creating generalizable skills, agents write \textbf{single-use scripts} with all values \textbf{hardcoded directly} into the code. The agent uses \texttt{exec\_script} to execute these scripts immediately, after which they are discarded. This eliminates both the abstraction overhead of designing reusable interfaces and the two-step save-execute workflow.

Table~\ref{tab:direct_exec} compares Base, Skill, and Direct Exec on a 48-task subset. For Claude-4.5-Sonnet, Direct Exec largely preserves success at 96\% while cutting tokens from 1.72M to 0.16M, and it reduces turns from 15.7 to 5.8 with tool calls from 14.7 to 4.8. Skill mode is less aggressive at 0.34M tokens and it drops success to 90\%. For GPT-5.2, Direct Exec achieves the largest savings from 1.18M to 0.06M tokens and reduces turns from 24.5 to 4.5, but success falls from 94\% to 85\%, while Skill keeps 90\% at 0.26M tokens. Direct Exec also has lower Exec at 68\% versus 97\% to 99\% in Skill mode, matching the fact that removing the agent loop removes recovery and adaptation. These results show Direct Exec as the efficiency upper bound when Skills transfer cleanly as standalone programs.
This advantage stems from two factors: (1) \textbf{reduced cognitive load}---the agent need not design generalizable parameter interfaces or anticipate future reuse scenarios; and (2) \textbf{simplified execution}---hardcoded values eliminate potential parameter binding errors that can occur in parameterized skill execution.

These results suggest that the generalization capability of Skills incurs a non-trivial overhead. When tasks are isolated and patterns are unlikely to be reused, Direct Exec Mode provides a more efficient alternative.

\subsection{Trajectory analysis}
\label{app:traj}


\definecolor{TrajPurple}{HTML}{4A148C}      
\definecolor{TrajPurpleLight}{HTML}{7B1FA2} 
\definecolor{TrajPurpleBg}{HTML}{F3E5F5}    
\definecolor{TrajGray}{HTML}{424242}        
\definecolor{TrajLightGray}{HTML}{F5F5F5}   

\newcommand{\TrajReset}{\setcounter{trajstep}{0}}

\newtcolorbox{trajectory}[1]{
  enhanced,
  breakable,
  colback=white,
  colframe=TrajPurple,
  boxrule=1pt,
  arc=3mm,
  left=4mm,
  right=4mm,
  top=3mm,
  bottom=3mm,
  fonttitle=\bfseries\normalsize,
  coltitle=white,
  colbacktitle=TrajPurple,
  title=#1,
  attach boxed title to top left={yshift=-2mm, xshift=-1mm},
  boxed title style={boxrule=0pt, arc=2mm, left=3mm, right=3mm}
}

\newtcolorbox{syscard}[1]{
  enhanced,
  colback=TrajLightGray,
  colframe=TrajLightGray,
  boxrule=0pt,
  arc=2mm,
  left=3mm,
  right=3mm,
  top=2mm,
  bottom=2mm,
  before skip=2mm,
  after skip=2mm,
  fontupper=\small,
  before upper={\textbf{#1}\par\smallskip}
}

\newtcolorbox{stepcard}{
  enhanced,
  colback=TrajPurpleBg,
  colframe=TrajPurpleBg,
  boxrule=0pt,
  arc=2mm,
  left=3mm,
  right=3mm,
  top=2mm,
  bottom=2mm,
  before skip=2mm,
  after skip=2mm,
  fontupper=\small,
  borderline west={2pt}{0pt}{TrajPurple}
}

\newtcolorbox{respcard}{
  enhanced,
  colback=TrajLightGray,
  colframe=TrajLightGray,
  boxrule=0pt,
  arc=2mm,
  left=3mm,
  right=3mm,
  top=2mm,
  bottom=2mm,
  before skip=1mm,
  after skip=2mm,
  fontupper=\small
}

\newtcolorbox{statsbox}[1]{
  enhanced,
  colback=TrajPurpleBg,
  colframe=TrajPurple,
  boxrule=0.8pt,
  arc=2mm,
  left=3mm,
  right=3mm,
  top=2mm,
  bottom=2mm,
  before skip=3mm,
  fontupper=\small,
  title={\small\bfseries #1},
  coltitle=TrajPurple,
  colbacktitle=white,
  attach boxed title to top left={yshift=-2mm, xshift=3mm},
  boxed title style={boxrule=0.8pt, colframe=TrajPurple, arc=1.5mm, left=2mm, right=2mm}
}

\newcommand{\tcode}[1]{\textit{#1}}

\newcommand{\stepnum}{\refstepcounter{trajstep}\textbf{Step \#\thetrajstep}}

\newcommand{\abbrev}[1]{{\color{TrajGray}\itshape\small\hspace{3mm}#1}}


\TrajReset
\begin{trajectory}{Traj A: Claude-4.5-Sonnet \quad\textit{\small cat-facts-collector/e2}}

\begin{syscard}{System}
You are an AI assistant with access to tools. Use save\_skill to create reusable workflows and execute\_skill to run them.
\end{syscard}

\begin{syscard}{User}
Collect information for 3 cat breeds (Persian, Siamese, Maine Coon) using breed\_profile, breed\_facts, and breed\_encyclopedia APIs. Save results to a JSON file.
\end{syscard}

\begin{stepcard}
\stepnum\quad Agent\par\smallskip
I'll process each breed systematically using direct API calls.\par
\textbf{call} \tcode{catfacts\_breed\_profile} \quad \tcode{Persian}
\end{stepcard}

\begin{respcard}
\{"origin": "Persia", "temperament": "Gentle", "life\_span": "12-17"...\}
\end{respcard}

\begin{stepcard}
\stepnum\quad Agent\par\smallskip
\textbf{call} \tcode{catfacts\_breed\_facts} \quad \tcode{Persian}
\end{stepcard}

\begin{respcard}
\{"fact\_1": "Persians are known for their long fur"...\}
\end{respcard}

\begin{stepcard}
\stepnum\quad Agent\par\smallskip
\textbf{call} \tcode{catfacts\_breed\_encyclopedia} \quad \tcode{Persian}
\end{stepcard}

\begin{respcard}
\{"description": "The Persian cat is a long-haired breed..."...\}
\end{respcard}

\begin{stepcard}
\stepnum\quad Agent\par\smallskip
Persian complete. Continuing with Siamese...
\end{stepcard}

\abbrev{Steps \#5--7: Same 3 API calls for Siamese and Maine Coon}

\begin{stepcard}
\textbf{Step \#8}\quad Agent\par\smallskip
\textbf{call} \tcode{write\_file} \quad \tcode{cat\_breeds\_report.json}
\end{stepcard}

\begin{respcard}
File written successfully.
\end{respcard}

\begin{stepcard}
\textbf{Step \#9}\quad Agent\par\smallskip
\textbf{call} \tcode{claim\_done} \quad Task completed successfully.
\end{stepcard}

\begin{statsbox}{Statistics --- Pass}
\textbf{Total Steps:} 9 \quad\quad \textbf{Total Tokens:} 762,052\par\smallskip
\hrule\smallskip
$\checkmark$ No skill created --- Direct API calls sufficient.
\end{statsbox}

\end{trajectory}

\vspace{4mm}


\TrajReset
\begin{trajectory}{Traj B: DeepSeek-V3.2 \quad\textit{\small cat-facts-collector/e2}}

\begin{syscard}{System}
You are an AI assistant with access to tools. Use save\_skill to create reusable workflows and execute\_skill to run them.
\end{syscard}

\begin{syscard}{User}
Collect information for 3 cat breeds (Persian, Siamese, Maine Coon) using breed\_profile, breed\_facts, and breed\_encyclopedia APIs. Save results to a JSON file.
\end{syscard}

\begin{stepcard}
\stepnum\quad Agent\par\smallskip
\textbf{call} \tcode{list\_directory} \quad \tcode{workspace}
\end{stepcard}

\begin{respcard}
[workspace contents]
\end{respcard}

\begin{stepcard}
\stepnum\quad Agent\par\smallskip
\textbf{call} \tcode{save\_skill} \quad \tcode{process\_cat\_breed}
\end{stepcard}

\begin{respcard}
Skill saved successfully.
\end{respcard}

\begin{stepcard}
\stepnum\quad Agent\par\smallskip
\textbf{call} \tcode{execute\_skill}\par
skill\_name: \tcode{process\_cat\_breed} \quad param: \tcode{Persian}
\end{stepcard}

\begin{respcard}
Skill executed.\par
$\triangleright$ Warning: missing breed\_facts, breed\_encyclopedia fields
\end{respcard}

\begin{stepcard}
\stepnum\quad Agent\par\smallskip
\textbf{call} \tcode{execute\_skill}\par
skill\_name: \tcode{process\_cat\_breed} \quad param: \tcode{Siamese}
\end{stepcard}

\begin{respcard}
Skill executed.\par
$\triangleright$ Warning: missing breed\_facts, breed\_encyclopedia fields
\end{respcard}

\begin{stepcard}
\stepnum\quad Agent\par\smallskip
\textbf{call} \tcode{execute\_skill}\par
skill\_name: \tcode{process\_cat\_breed} \quad param: \tcode{Maine Coon}
\end{stepcard}

\begin{respcard}
Skill executed.\par
$\triangleright$ Warning: missing breed\_facts, breed\_encyclopedia fields
\end{respcard}

\begin{stepcard}
\stepnum\quad Agent\par\smallskip
\textbf{call} \tcode{write\_file} \quad \tcode{partial\_results.json}
\end{stepcard}

\begin{respcard}
File written.
\end{respcard}

\begin{stepcard}
\stepnum\quad Agent\par\smallskip
\textbf{call} \tcode{file\_write\_json\_chunk} \quad \tcode{persian\_data}
\end{stepcard}

\begin{respcard}
Chunk written.
\end{respcard}

\abbrev{Steps \#8--14: More repair operations (write\_json\_chunk $\times$4, read\_file, file\_append)}

\begin{stepcard}
\textbf{Step \#15}\quad Agent\par\smallskip
\textbf{call} \tcode{claim\_done} \quad Task completed.
\end{stepcard}

\begin{statsbox}{Statistics --- Pass}
\textbf{Total Steps:} 15 \quad\quad \textbf{Total Tokens:} 1,507,830\par\smallskip
\hrule\smallskip
$\times$ Skill created unnecessarily --- 8 extra repair calls needed.
\end{statsbox}

\end{trajectory}

\vspace{4mm}


\TrajReset
\begin{trajectory}{Traj C: Claude-4.5-Sonnet \quad\textit{\small cocktail-menu-generator/h1}}

\begin{syscard}{System}
You are an AI assistant with access to tools. Use save\_skill to create reusable workflows and execute\_skill to run them.
\end{syscard}

\begin{syscard}{User}
Create a cocktail menu for 5 cocktails (Margarita, Mojito, Old Fashioned, Martini, Negroni) using 5 APIs each. Save to JSON.
\end{syscard}

\begin{stepcard}
\stepnum\quad Agent\par\smallskip
25 calls total. I'll create a reusable skill.\par
\textbf{call} \tcode{save\_skill} \quad \tcode{process\_cocktail\_complete}
\end{stepcard}

\begin{respcard}
Skill saved successfully.
\end{respcard}

\begin{stepcard}
\stepnum\quad Agent\par\smallskip
Now processing all 5 cocktails...\par
\textbf{call} \tcode{execute\_skill}\par
skill\_name: \tcode{process\_cocktail\_complete} \quad param: \tcode{Margarita}
\end{stepcard}

\begin{respcard}
\{search, details, ingredient, category, glass\} $\rightarrow$ all fields complete
\end{respcard}

\abbrev{Steps \#3--6: execute\_skill for Mojito, Old Fashioned, Martini, Negroni $\rightarrow$ all success}

\begin{stepcard}
\textbf{Step \#7}\quad Agent\par\smallskip
\textbf{call} \tcode{write\_file} \quad \tcode{cocktail\_menu.json}
\end{stepcard}

\begin{respcard}
File written successfully.
\end{respcard}

\begin{stepcard}
\textbf{Step \#8}\quad Agent\par\smallskip
\textbf{call} \tcode{claim\_done} \quad Cocktail menu saved. Task completed.
\end{stepcard}

\begin{statsbox}{Statistics --- Pass}
\textbf{Total Steps:} 8 \quad\quad \textbf{Total Tokens:} 213,222\par\smallskip
\hrule\smallskip
$\checkmark$ 1 save $\rightarrow$ 5 executes --- Skill correct on first try.
\end{statsbox}

\end{trajectory}

\vspace{4mm}


\TrajReset
\begin{trajectory}{Traj D: DeepSeek-V3.2 \quad\textit{\small cocktail-menu-generator/h1}}

\begin{syscard}{System}
You are an AI assistant with access to tools. Use save\_skill to create reusable workflows and execute\_skill to run them.
\end{syscard}

\begin{syscard}{User}
Create a cocktail menu for 5 cocktails (Margarita, Mojito, Old Fashioned, Martini, Negroni) using 5 APIs each. Save to JSON.
\end{syscard}

\begin{stepcard}
\stepnum\quad Agent\par\smallskip
\textbf{call} \tcode{save\_skill} \quad \tcode{process\_cocktail}
\end{stepcard}

\begin{respcard}
Skill save failed.\par
$\triangleright$ SyntaxError: unexpected token `\}' at line 8
\end{respcard}

\begin{stepcard}
\stepnum\quad Agent\par\smallskip
Fix syntax error and retry...\par
\textbf{call} \tcode{save\_skill} \quad \tcode{process\_cocktail\_v2}
\end{stepcard}

\begin{respcard}
Skill saved.
\end{respcard}

\begin{stepcard}
\stepnum\quad Agent\par\smallskip
\textbf{call} \tcode{execute\_skill}\par
skill\_name: \tcode{process\_cocktail\_v2} \quad param: \tcode{Margarita}
\end{stepcard}

\begin{respcard}
Execution failed.\par
$\triangleright$ SyntaxError: `return' is invalid outside function (line 12)
\end{respcard}

\begin{stepcard}
\stepnum\quad Agent\par\smallskip
Skill failed. Fallback to direct API...\par
\textbf{call} \tcode{cocktail\_search} \quad \tcode{Margarita}
\end{stepcard}

\begin{respcard}
\{"id": "11007", "name": "Margarita"...\}
\end{respcard}

\begin{stepcard}
\stepnum\quad Agent\par\smallskip
\textbf{call} \tcode{cocktail\_details} \quad \tcode{11007}
\end{stepcard}

\begin{respcard}
\{"ingredients": ["Tequila", "Lime"...]...\}
\end{respcard}

\begin{stepcard}
\stepnum\quad Agent\par\smallskip
\textbf{call} \tcode{cocktail\_by\_ingredient} \quad \tcode{Tequila}
\end{stepcard}

\begin{respcard}
\{"cocktails": [...]...\}
\end{respcard}

\begin{stepcard}
\stepnum\quad Agent\par\smallskip
\textbf{call} \tcode{cocktail\_by\_ingredient} \quad \tcode{Lime}
\end{stepcard}

\begin{respcard}
\{"cocktails": [...]...\}
\end{respcard}

\begin{stepcard}
\stepnum\quad Agent\par\smallskip
\textbf{call} \tcode{save\_skill} \quad \tcode{process\_cocktail\_v3}
\end{stepcard}

\begin{respcard}
Skill saved.
\end{respcard}

\abbrev{Steps \#9--18: More execute\_skill calls + manual cleanup}

\begin{stepcard}
\textbf{Step \#19}\quad Agent\par\smallskip
\textbf{call} \tcode{claim\_done} \quad Task completed after multiple retries.
\end{stepcard}

\begin{statsbox}{Statistics --- Fail}
\textbf{Total Steps:} 19 \quad\quad \textbf{Total Tokens:} 1,141,166\par\smallskip
\hrule\smallskip
$\times$ 3 saves + fallback --- Skill quality poor, task failed.
\end{statsbox}

\end{trajectory}


We present representative trajectories from our experiments to illustrate the qualitative differences in how models approach skill creation and reuse. The above shows four trajectories: two from an easy task (\texttt{cat-facts-collector/e2}) and two from a hard task (\texttt{cocktail-menu-generator/h1}), comparing Claude-4.5-Sonnet and DeepSeek-V3.2.

\paragraph{Behavioral Divergence.}
A fundamental distinction emerges in how models decide \emph{whether} to create skills. Claude exhibits efficiency-maximizing behavior: it autonomously evaluates whether the abstraction overhead is justified before committing to skill creation. In Trajectory A, Claude identifies that the easy task (9 API calls for 3 cat breeds) does not warrant skill abstraction and proceeds with direct calls, completing in 34 steps. In Trajectory C, facing a harder task (15 API calls for 5 cocktails), Claude creates a single skill that executes correctly 5 times with zero errors. In contrast, DeepSeek follows the system prompt more literally, attempting skill creation regardless of task complexity. In Trajectory B, it creates \texttt{process\_cat\_breed} for the same easy task despite minimal reuse benefit, and in Trajectory D, it persists through three failed skill creation attempts before abandoning the approach entirely.

\paragraph{Skill Creation Failures.}
DeepSeek's skill creation attempts reveal systematic issues. In Trajectory B, the created skill \texttt{process\_cat\_breed} is incomplete---its output schema omits \texttt{breed\_facts} and \texttt{breed\_encyclopedia} fields, requiring 8 additional repair operations. In Trajectory D, DeepSeek attempts skill creation three times (\texttt{process\_cocktail}, \texttt{process\_cocktail\_v2}, \texttt{process\_cocktail\_v3}), each failing with syntax errors such as ``unexpected token'' and ``return is invalid outside function.'' These errors indicate that DeepSeek treats skill creation as template expansion rather than program synthesis.

\paragraph{Skill Execution Failures.}
Even when skills are successfully saved, execution failures reveal deeper issues. In Trajectory B, all three \texttt{execute\_skill} calls produce incomplete results with warnings about missing fields. The skill's internal logic failed to properly chain the three required API calls. In Trajectory D, the \texttt{execute\_skill} call fails immediately with a runtime error, forcing the agent to fall back to manual API calls and ultimately failing the task.

\paragraph{Implications.}
These findings suggest that effective tool composition requires not just the \emph{ability} to create and execute skills, but the \emph{judgment} to know when abstraction is beneficial. The 5.3$\times$ token savings achieved by Claude in the hard task (213K vs. 1.14M tokens) compared to DeepSeek demonstrates that understanding-driven skill use leads to both higher success rates and greater efficiency.

\section{Prompt Templates}
\label{app:prompts}

This section presents the prompt templates used in our experiments, including the system prompt for skill-enabled modes and representative task prompts across different difficulty levels.

\subsection{System Prompt for Skill Reuse}
\label{app:system_prompt}

In skill mode, agents receive an augmented system prompt that introduces the skill abstraction mechanism. The prompt provides: (1) available skill tools (\texttt{save\_skill} and \texttt{execute\_skill}); (2) guidelines for when to create skills; (3) script authoring rules; and (4) a concrete example demonstrating the skill creation and execution workflow.

The key design principle is \emph{minimal intervention}: rather than prescribing when agents should use skills, we provide the capability and let agents autonomously decide based on task structure. This enables fair comparison between skill-enabled and baseline modes, as the core task instructions remain identical.

\begin{SystemPromptBox}{System Prompt: Skill Reuse Mode}

\SPLabel{Skill Tools:}
You have access to skill cache tools to save and execute reusable scripts:
\begin{itemize}
  \setlength{\itemsep}{2pt}
  \setlength{\parskip}{0pt}
  \setlength{\parsep}{0pt}
  \item \texttt{save\_skill} --- Save an executable script as a reusable skill
  \item \texttt{execute\_skill} --- Execute a saved skill with different arguments
\end{itemize}

\SPDivider

\SPLabel{When to Use:}
For repetitive operations (processing multiple items, files, etc.), create a skill to encapsulate the workflow, then execute it for all items. You can create skills based on tool schemas without calling the tool first---especially efficient when tools return large data.

\SPDivider

\SPLabel{Script Rules:}
\begin{enumerate}
  \setlength{\itemsep}{2pt}
  \setlength{\parskip}{0pt}
  \setlength{\parsep}{0pt}
  \item Use \texttt{call\_tool()} for ALL tool calls: \texttt{call\_tool('tool\_name', arg1=val1, ...)}
  \item \texttt{call\_tool()} returns DIRECT result---use it directly without \texttt{.get("result")} wrapper
  \item MUST set \texttt{result} variable---this is what gets returned from \texttt{execute\_skill}
  \item Modules available: \texttt{re}, \texttt{json}, \texttt{os} are pre-imported
  \item No recursion: Cannot call skill tools within skills
\end{enumerate}

\SPDivider

\SPLabel{Example:}
{\small
\begin{verbatim}
save_skill({
  "skill_name": "analyze_project",
  "script_code": ...
})

execute_skill({
    "skill_name": "analyze_project"
    "args": {"path": "org/repo1"}}
\end{verbatim}
}

\SPDivider

\SPLabel{Best Practices:}
\begin{itemize}
  \setlength{\itemsep}{2pt}
  \setlength{\parskip}{0pt}
  \setlength{\parsep}{0pt}
  \item \textbf{Token Efficiency:} Extract only fields needed for final output
  \item \textbf{Maximize ROI:} Create skill early, execute for ALL items (beneficial when $N \geq 3$)
  \item \textbf{Fallback:} If skill fails 2--3 times, process items directly
\end{itemize}

\end{SystemPromptBox}

\subsection{Task Prompt Examples}
\label{app:task_prompts}

Task prompts describe the objective, required outputs, and available domain-specific tools. We present three representative examples from our scaled task suite, spanning easy (E), medium (M), and hard (H) difficulty levels. The scaling follows a systematic pattern: easy tasks involve $3 \times 3 = 9$ API calls, medium tasks involve $4 \times 4 = 16$ calls, and hard tasks involve $5 \times 5 = 25$ calls.

Each prompt specifies:
\begin{itemize}
    \item \textbf{Objective}: The data collection or analysis goal
    \item \textbf{Output format}: JSON schema for structured results
    \item \textbf{Available tools}: Domain-specific APIs (prefixes removed for clarity)
    \item \textbf{Scale}: Number of subtasks and API calls per subtask
\end{itemize}

Note that skill-related tools (\texttt{save\_skill}, \texttt{execute\_skill}) are \emph{not} mentioned in task prompts---they are injected via the system prompt only in skill-enabled modes. This ensures that baseline (Normal) mode and skill-enabled modes receive identical task instructions.



\begin{TaskPromptBox}{cat-facts-collector/e1 ~~{\small[Easy]}}

\TPLabel{Prompt:}
Create encyclopedia entries for \textbf{3 cat breeds} (Persian, Siamese, Maine Coon) using 3 API endpoints per breed. For each breed, collect: (1) \textit{Breed Profile} --- basic info and characteristics; (2) \textit{Country Relatives} --- breeds from same country; (3) \textit{Coat Family} --- breeds with similar coat. Compile a summary with statistics across all breeds and save results to cat\_encyclopedia.json.

\TPDivider

\TPLabel{Available tools:}
\begin{itemize}
  \setlength{\itemsep}{4pt}
  \setlength{\parskip}{0pt}
  \setlength{\parsep}{0pt}
  \item breed\_profile(breed\_name)\\
        {\small Get breed info and characteristics}
  \item breed\_relatives(country)\\
        {\small List breeds from same country}
  \item breed\_coat\_family(coat\_type)\\
        {\small List breeds with similar coat}
  \item write\_file(path, content)\\
        {\small Save JSON output}
  \item claim\_done(status)\\
        {\small Signal task completion}
\end{itemize}

\TPDivider

{\small\color{gray} \textbf{Scale:} 3 subtasks $\times$ 3 API calls = 9 total calls}

\end{TaskPromptBox}

\vspace{4mm}


\begin{TaskPromptBox}{cocktail-menu-generator/m1 ~~{\small[Medium]}}

\TPLabel{Prompt:}
Create a cocktail menu for \textbf{4 classic cocktails} (Margarita, Mojito, Old Fashioned, Martini) using 4 API endpoints per cocktail. For each cocktail, collect: (1) \textit{Search} --- find cocktail by name; (2) \textit{Details} --- full recipe and instructions; (3) \textit{By Ingredient} --- list cocktails using main ingredient; (4) \textit{By Category} --- list cocktails in same category. Calculate complexity rating (Easy/Medium/Complex based on ingredient count) and estimated prep time. Save results to cocktail\_menu.json.

\TPDivider

\TPLabel{Available tools:}
\begin{itemize}
  \setlength{\itemsep}{4pt}
  \setlength{\parskip}{0pt}
  \setlength{\parsep}{0pt}
  \item search(name)\\
        {\small Search cocktail by name}
  \item details(id)\\
        {\small Get full recipe and instructions}
  \item by\_ingredient(ingredient)\\
        {\small List cocktails with ingredient}
  \item by\_category(category)\\
        {\small List cocktails in category}
  \item write\_file(path, content)\\
        {\small Save JSON output}
  \item claim\_done(status)\\
        {\small Signal task completion}
\end{itemize}

\TPDivider

{\small\color{gray} \textbf{Scale:} 4 subtasks $\times$ 4 API calls = 16 total calls}

\end{TaskPromptBox}

\vspace{4mm}


\begin{TaskPromptBox}{gitlab-deep-analysis/h1 ~~{\small[Hard]}}

\TPLabel{Prompt:}
Perform a comprehensive analysis of \textbf{5 GitLab repositories} (gitlab-runner, gitaly, gitlab-pages, gitlab-shell, cli). For each project, collect: (1) \textit{Project Info} --- stars, forks, description; (2) \textit{Contributors} --- top 5 by commit count; (3) \textit{Recent Commits} --- last 20 commits with authors; (4) \textit{Branches} --- all branches with protection status; (5) \textit{Issues} --- open count and recent titles. Calculate activity score (0--100) based on commits (40\%), contributors (30\%), issues (20\%), branches (10\%). Determine health status: \textit{healthy} ($\geq$70), \textit{moderate} (40--70), \textit{inactive} ($<$40). Save results to gitlab\_analysis\_results.json.

\TPDivider

\TPLabel{Available tools:}
\begin{itemize}
  \setlength{\itemsep}{4pt}
  \setlength{\parskip}{0pt}
  \setlength{\parsep}{0pt}
  \item get\_project\_info(project\_path)\\
        {\small Get project details (stars, forks, description)}
  \item get\_contributors(project\_path)\\
        {\small Get contributor list}
  \item get\_commits(project\_path, limit)\\
        {\small Get commit history}
  \item get\_branches(project\_path)\\
        {\small Get branch information}
  \item get\_issues(project\_path)\\
        {\small Get issue list}
  \item write\_file(path, content)\\
        {\small Save JSON output}
  \item claim\_done(status)\\
        {\small Signal task completion}
\end{itemize}

\TPDivider

{\small\color{gray} \textbf{Scale:} 5 subtasks $\times$ 5 API calls = 25 total calls}

\end{TaskPromptBox}

\end{document}